\def\final{0}
\documentclass[12pt]{article}

\usepackage{expl3}

\usepackage{algorithm, algorithmic}
\usepackage{amsmath, amssymb, amsthm, graphicx}
\usepackage{framed}
\usepackage{fullpage}
\usepackage{verbatim}
\usepackage{xcolor}

\usepackage{tcolorbox}

\usepackage[colorlinks]{hyperref}
\usepackage[capitalize,nameinlink]{cleveref}
\hypersetup{colorlinks={true},linkcolor={magenta},citecolor=teal}

\usepackage[shortlabels]{enumitem}

\usepackage{setspace}
\usepackage{geometry}
\usepackage{epstopdf}     
\ifnum\final=0
\newcommand{\mynote}[1]{\marginpar{\tiny \sf #1}}
\else
\newcommand{\mynote}[1]{}
\fi

\newcommand{\ignore}[1]{}

\newcommand{\F}{{\cal F}}

\newcommand{\Q}{{\cal Q}}
\newcommand{\X}{{\cal X}}

\newcommand{\R}{{\mathbb R}}

\newcommand{\tv}{\mathsf{TV}}
\newcommand{\N}{\mathbb{N}}

\newcommand{\Ex}{\mathbb{E}}

\newcommand{\opt}{\mathsf{opt}}

\newcommand\bH{\mathbb{H}}

\newcommand{\cE}{\mathcal{E}}

\newcommand{\cH}{\mathcal{H}}

\newcommand{\cP}{\mathcal{P}}
\newcommand{\cQ}{\mathcal{Q}}

\newcommand{\cX}{\mathcal{X}}

\newcommand{\poly}{\mathrm{poly}}

\newcommand{\unitint}{[0,1]}

\newcommand{\eps}{\varepsilon}
\newcommand{\kl}{\mathsf{KL}}
\newcommand{\player}{\mathsf{player}}
\newcommand{\adv}{\mathsf{adv}}
\newcommand{\argmax}{\operatorname{argmax}}
\newcommand{\argmin}{\operatorname{argmin}}

\newtheorem{theorem}{Theorem}
\newtheorem*{theorem*}{Theorem}
\newtheorem{lemma}[theorem]{Lemma}

\newtheorem{claim}[theorem]{Claim}

\newtheorem{proposition}[theorem]{Proposition}

\theoremstyle{definition}

\newcommand{\ga}{\gamma}

\title{{Statistically Near-Optimal Hypothesis Selection}}
\author{
Olivier Bousquet\thanks{Google Brain, Z{\"u}rich. {\tt obousquet@google.com.}}
\and Mark Braverman\thanks{Department of Computer Science, Princeton University. {\tt mbraverm@princeton.edu.}}
\and Klim Efremenko\thanks{Department of Computer Science, Ben Gurion University. {\tt klimefrem@gmail.com.}}
\and Gillat Kol\thanks{Department of Computer Science, Princeton University. {\tt gillat.kol@gmail.com.}}
\and Shay Moran\thanks{Department of Mathematics, Technion and Google Research. {\tt  smoran@technion.ac.il.}}
}

\date{}
\begin{document}

\maketitle
\thispagestyle{empty}

\abstract{
{\em Hypothesis Selection} is a fundamental distribution learning problem where given a comparator-class $\Q=\{q_1,\ldots, q_n\}$ of distributions, and a sampling access to an unknown target distribution $p$, the goal is to output a distribution $q$ such that $\tv(p,q)$ is close to $\opt$, where $\opt = \min_i\{\tv(p,q_i)\}$ and $\tv(\cdot, \cdot)$ denotes the total-variation distance. Despite the fact that this problem has been studied since the 19th century, 
its complexity in terms of basic resources, such as number of samples and approximation guarantees, remains unsettled (this is discussed, {\em e.g.},  in the charming book by Devroye and Lugosi `00). This is in stark contrast with other (younger) learning settings, such as PAC learning, for which these complexities are well understood.

We derive an {\em optimal $2$-approximation} learning strategy for the Hypothesis Selection problem, outputting $q$ such that $\tv(p,q) \leq2 \cdot \opt + \eps$, with a {\em (nearly) optimal sample complexity} of~$\tilde O(\log n/\eps^2)$. This is the first algorithm that simultaneously achieves the best approximation factor and  sample complexity: previously, Bousquet, Kane, and Moran ({\it COLT `19}) gave a learner achieving the optimal $2$-approximation, 
but with an exponentially worse sample complexity of $\tilde O(\sqrt{n}/\eps^{2.5})$,
and Yatracos~({\it Annals of Statistics `85}) gave a learner with optimal sample complexity of $O(\log n /\eps^2)$ but with a sub-optimal approximation factor of $3$.

We mention that many works in the {\em Density Estimation} ({\em a.k.a.}, {\em Distribution Learning}) literature use Hypothesis Selection as a black box subroutine. Our result therefore implies an improvement on the approximation factors obtained by these works, while keeping their sample complexity intact. For example, our result improves the approximation factor of the algorithm of Ashtiani, Ben-David, Harvey, Liaw, and Mehrabian~({\it JACM '20}) for  {\it agnostic learning of mixtures of gaussians} from $9$ to~$6$, while maintaining its nearly-tight sample complexity.

\newpage

\section{Introduction}

{\em Hypothesis selection} is a fundamental task in statistics, where a {\em learner} is getting a sample access to an {\em unknown} distribution $p$ on some, possibly infinite, domain~$\cX$, and wishes to output a distribution~$q$ that is ``close'' to~$p$. The problem was studied extensively over the last century and found many applications, most notably, in machine learning.

In this paper we study the hypothesis selection problem in the {\em agnostic} setting, where we assume a fixed finite\footnote{See discussion of the infinite case at the end of this section.} class $\Q$ of reference distributions which is known to the learner, and which may or may not contain $p$\footnote{
The setting where $p$ is assumed to be in $\Q$ is called the {\em realizable} setting.}.
The goal of the learner is to output a distribution $q$ that is at least as close to~$p$ as any of the distributions in $\Q$ in {\em total variation} distance (denoted here $\tv(\cdot, \cdot)$).

The statistical performance of a learner is measured using two parameters, denoted $\alpha$ and $m=m(n,\eps,\delta)$, where $\alpha$ is the {\em approximation factor} of the algorithm and $m$ is its {\em sample complexity}.
Specifically, we say that a class of distributions $\Q = \{q_1,\ldots,q_n\}$ is {\it $\alpha$-learnable with sample complexity $m(n,\eps,\delta)$} if there is a {(possibly randomized)} learner such that for every $\eps,\delta>0$ and every target distribution~$p$, upon receiving $m(n,\eps,\delta)$ random samples from~$p$, the learner outputs a distribution $q$ satisfying $\tv(p,q) \leq \alpha \cdot \min_{i \in [n]}\{\tv(p,q_i)\} + \eps$ with probability at least $1-\delta$. For the discussion below, we think of $\delta$ as a small constant.

\paragraph{How good can a learner be?} A-priori, it is not even clear that every class $\Q$ is learnable with finite sample complexity. Consider the following natural algorithm for 
hypothesis selection: estimate $\tv(q_i,p)$ for every $q_i \in Q$ and output the $q_i$ that minimizes this quantity. While this algorithm clearly works (and even achieves an approximation factor of $\alpha = 1$), estimating $\tv(q_i,p)$ for any $q_i$ requires $\tilde{\Omega}(\lvert \X\rvert)$ samples from $p$ (see, {\em e.g.},~\cite{Jiao18minimax}). Thus, if the domain~$\cX$ is infinite  (say~$\X=\mathbb{R}$), the sample complexity of this algorithm is not even finite. However, perhaps surprisingly, despite the impossibility of estimating the distance of $p$ from even one of the distributions $q_i$, one can still find an approximate minimizer of the distances (even when $\X$ is infinite!). 

What are the smallest $\alpha$ and $m$ for which any given class of distributions~$\Q$ of size $n$ is $\alpha$-learnable with sample complexity $m$?
A seminal work by Yatracos \cite{yatracos85} (also see \cite{devroye1996universally, devroye1997nonasymptotic, Devroye01combinatorial}) shows that any reference class $Q$ of size $n$ is $3$-learnable with sample complexity $O(\log n/\eps^2)$. 
For the case of $n=2$, Mahalanabis and Stefankovic \cite{Mahalanabis08density} improve the approximation factor, constructing a $2$-learner. 
This was extended by the recent work of Bousquet, Kane, and Moran~\cite{BKM19} to give a $2$-approximation for any finite $n$, using a very different scheme. A matching lower bound of $2$ on the approximation factor follows from the work of~\cite{Chan14histograms}.

Although the work of~\cite{BKM19} obtains the optimal approximation factor for the agnostic hypothesis selection problem, the sample complexity of their scheme is  $\tilde O(\sqrt{n}/\eps^{2.5})$, which is exponential in the sample complexity of Yatracos's algorithm\footnote{We note that~\cite{BKM19} also provide $\poly(\log\lvert \X\rvert, \log n, \eps^{-1})$ sample complexity bounds, which can be better than their general $\tilde O(\sqrt{n}/\eps^{2.5})$ bound for finite domains $\cX$.}.
Deriving optimal learners with efficient sample complexity is left as the main open problem in their work. In this paper, we give a novel $2$-learner with (near) optimal sample complexity, getting the best of both worlds.

\paragraph{Density Estimation.} 
Hypothesis selection, and, in particular, Yatracos's algorithm, found applications beyond learning finite classes. Specifically, it is used as a basic subroutine in density estimation tasks where the goal is to learn an infinite class of distributions, in the realizable or agnostic setting\footnote{In fact, learning infinite classes was a part of Yatracos's original motivation.}. A popular method, where the reference class $\Q$ may be infinite, is the {\em cover method} ({\em a.k.a.}\ the skeleton method). In this method, one ``covers'' the class $\Q$ by a finite $\alpha$-cover; that is, a subclass $\Q'\subseteq\Q$ of distributions such that for every $q \in \Q$ there exists $q' \in \Q'$ with $\tv(q,q') \leq \alpha$. Often times it is the case that even if $\Q$ is infinite, a finite $\eps$-net $\Q'$ exists, and Yatracos's agnostic learning algorithm can be applied on $\Q'$ (see \cite{Devroye01combinatorial,Diakonikolas16survey} and references within for many such examples).

While the minimal possible size of such a cover $\Q'$ is often exponential in the natural parameters of the class $\Q$\footnote{One easy example of an exponential cover is when $\Q$ is the set of all convex combinations of $k$ fixed distributions $p_1,\ldots,p_k$, {\em i.e.}, $\Q = \{ \sum_{i \in [k]} \beta_i p_i :\; \sum_{i \in [k]} \beta_i = 1, \beta_i\geq 0 \}$. The set $\Q = \{ \sum_{i \in [k]} \frac{r_i}{\ell} \cdot p_i :\; r_i \in \N \cup \{0\}, \; \ell = \lceil \frac{k}{\alpha} \rceil, \; \sum_{i \in [k]}\frac{r_i}{\ell} = 1 \}$ is a cover of $\Q$ of exponential size (in $k$). Sub-exponential covers are not possible in this case. See Chapter $7.4$ in \cite{Devroye01combinatorial} for this example, and the rest of Chapter $7$ for more such examples.}, because Yatracos's algorithm has poly-logarithmic sample complexity, the obtained density estimation algorithm has a polynomial sample complexity.
Since many density estimation results follow the cover method, or other related methods\footnote{Another such method is the recent {sample compression} method by \cite{Ashtiani18mixtures}, used to obtain improved density algorithms for the mixtures of Gaussians problem.} that use Yatracos's algorithm as a black box, our algorithm can imply an improvement for all of these results. (We mention a couple of such examples below, in \Cref{sec:related}).

We note that in the realizable setting for density estimation, where the distribution $p$ we wish to learn is in the infinite class $\Q$ of distributions we are considering (that is, $\opt = 0$), one can typically get a better approximation factor by taking a finer cover (smaller $\alpha$). By taking an $\alpha$-cover of $\Q$, the above method results in a distribution $q$ with $\tv(p,q) \leq \alpha + 3\opt = \alpha$. However, in the agnostic setting, even if we take a very small $\alpha$, the resulting $\tv(p,q)$ may not be small as it is dominated by $3 \opt$. By using the result of this paper in lieu of Yatracos's learning algorithm, this distance can be made $2 \opt$. 

\subsection{Our Results} \label{sec:results}

We design a $2$-learner for the agnostic hypothesis selection problem with sample complexity whose dependence on both $n$ and $\eps$ is (near) optimal.
\begin{theorem} \label{thm:main}
Let $\Q$ be a finite class of distributions and let $n=\lvert \Q\rvert$.
Then, $\Q$ is $2$-learnable with sample complexity\footnote{We use the standard notation that $f(n) = \tilde{O}(h(n_1,\ldots,n_t))$ if there exists $k \in \mathbb{N}$ such that $f(n_1,\ldots,n_t) = O(h(n_1,\ldots,n_t) \log^k(h(n_1,\ldots,n_t)))$.} 
$m(n,\eps,\delta) = \tilde{O} \left( (\log n \cdot \min(\log n, \log (1/\delta))+\log (1/\delta)) / \eps^2 \right)$.
In particular, for constant $\delta > 0$,
\[m(n,\eps,\delta) = \tilde{O} \left( \frac{\log n}{\eps^2}\right).\]
\end{theorem}

Our learner in \autoref{thm:main} is {\em deterministic}, and, as in the case for \cite{BKM19}, it only makes {\em statistical queries}. That is, our learner can be implemented in the restricted model where instead of getting random samples from $p$, the learner has access to an oracle that on a query $(f,\eps)$ answers by a value in $\Ex_{x \sim p}[f(x)] \pm \eps$ (or, equivalently, on a query $(F,\eps)$, where $F$ is a set, answers by $p(F) \pm \eps$). Furthermore, our algorithm consists of only $\tilde{O}(\log n/\eps^2)$ such rounds of queries, whereas the algorithm \cite{BKM19} consists of $O(n/\eps)$ such rounds. 
}

\subsection{Our Technique} \label{sec:intro-technique}

\subsubsection{The Cutting-With-Margin Game} \label{sec:intro-game}

To prove \autoref{thm:main}, we reduce the hypothesis selection problem to solving a geometric game we call the ``{\em cutting-with-margin}'' game. This game is between a player and an adversary and it is played over a convex body $\cH \subseteq \Delta_n$ known to both parties, where $\Delta_n$ denotes the simplex of $n$-dimensional probability vectors\footnote{{\em I.e.}, $\Delta_n := \bigl\{h\in \R^n : \sum_{i \in [n]} h_i = 1, ~ (\forall i): h_i \geq 0\bigr\}.$}. In every round of the game, the player selects a point $h \in \cH$ and adversary updates the set $\cH$ to a new convex set by ``cutting out'' a part of $\cH$ that contains the $\ell_1$ ball of radius $\eps$ around $h$. The game ends when the set $\cH$ is empty.

We first show that any strategy for the player which ensures that the game ends in at most $r$ rounds implies a $2$-learner for the hypothesis selection problem with sample complexity $\tilde{O}(r \log n / \eps^2)$ (this is because the implementation of each round requires $n$ statistical queries that should be approximated to within $O(\eps)$). 
We then give an information-theoretic argument showing that the game is solvable in $r = \tilde{O}\left({\log(n)}/{\eps^2}\right)$ rounds, implying a hypothesis selection algorithm with $\tilde{O}\left({\log^2(n)}/{\eps^4}\right)$ samples. Our player's strategy views each point $h \in \cH \subseteq \Delta_n$ as a distribution and takes the point $h \in \cH$ that {\em maximizes the entropy} function. 

Even though the cutting-with-margin game serves as a technical tool in this work, this simple game
may also be of independent interest, and it is natural to study it for different norms (other than the $\ell_1$ norm considered in this paper). In a sense, this game is a dual perspective on the geometric approach taken by \cite{BKM19} (see \Cref{sec:overview}). Nevertheless, it is the move to this dual perspective that allowed us to use the above maximum-entropy-based strategy. While entropy-based strategies are widely used in {online} optimization (see \Cref{sec:related}), we find the fact that such a strategy is helpful for making progress in this abstract statistical problem of hypothesis selection, to be curious. {We hope that this connection will inspire more collaboration between the optimization and the statistical learning communities.}


\subsubsection{Achieving Optimal Sample Complexity} 

Our solution for the cutting-with-margin game yields a hypothesis selection algorithm with sample complexity polynomial in $\log n /\eps$, but still sub-optimal. While reducing the sample complexity of this algorithm and achieving a near optimal complexity of $\tilde{O}(\log n/\eps^2)$ requires quite a bit of effort (in fact, it is the main technical contribution of this paper), we believe that it makes our algorithm more applicable (in the sense that it can replace Yatracos's algorithm, without compromising the sample complexity).

To this end, at a very high level, we consider a ``dynamic'' cutting-with-margin game that allows the cutting of $\ell_1$ balls of different diameters, 
and we give a ``{\em win-win}''-style strategy, where in rounds where we use more samples  the diameter of the ball we cut is larger (see \Cref{sec:optoverview}). Thus, the player either makes a lot of progress towards the goal or uses few samples.

A detailed overview of our techniques can be found in \Cref{sec:overview}.

\paragraph{{Adaptive data analysis.}} 
As explained in \Cref{sec:overview}, the (``primal'') geometric approach of \cite{BKM19} results in a hypothesis selection algorithm that makes $O(n^2/\eps)$ statistical queries, where each should be approximated to within $O(\eps)$. Had all these queries been submitted together, the standard combination of Chernoff and union bound would imply a logarithmic sample complexity. However, their algorithm submits these queries {\it adaptively}, in $O(n/\eps)$ rounds, where in each round~$n$ queries are submitted. Thus, naively, each of the rounds will require $\tilde O(\log n / \eps^2)$ fresh samples for the total sample complexity of $\tilde{O}(n / \eps^3)$. Their improved stated sample complexity of $\tilde O(\sqrt{n}/\eps^{2.5})$ is made possible by importing clever tools from {\it Adaptive Data Analysis}.

Given the above, a natural question is whether similar ``off-the-shelf'' Adaptive Data Analysis tools can be used to convert the hypothesis selection algorithm obtained in \Cref{sec:intro-game} from our solution of the cutting-with-margin game, to a sample optimal one. (Recall that this protocol consists of $\tilde{O}(\log n/\eps^2)$ rounds and makes $n$ statistical queries in each round). Unfortunately, we were unable to apply these tools to get a significant quantitative improvements, as these tools are mostly geared toward cases where there are many rounds of adaptivity, while in our algorithm, the number of rounds $\tilde{O}(\log n/\eps^2)$ is much smaller than the number of queries $n$ made in every round (see, {\em e.g.}, \cite{Dwork15Preserving}). Instead, as described above, we use a more direct solution and {tune the number of samples we use for each query {\em adaptively}, by monitoring (and verifying) the progress of the algorithm.} 

It will be interesting to explore whether our technique can be extended to more general protocols in adaptive data analysis.

\subsection{Additional Discussion of The Model}

In this work, we give an {\em improper} algorithm for the {\em finite} {\em agnostic} hypothesis selection problem under the {\em total variation distance}. 
 We next explain the modeling choices we have made:

\paragraph{The finite agnostic setting.}
We consider the finite agnostic setting;
clearly, an algorithm in this setting applies in the realizable setting as well.
In addition, as discussed above, hypothesis selection in the finite agnostic setting is often used as a building block in the infinite (agnostic and realizable) settings ({\em i.e.},\ in density estimation).



\paragraph{Total variation distance.} The total variation distance is used by numerous prior works in the field, and is a  natural choice for our study for several reasons: 
firstly, solving the hypothesis selection problem for the total variation distance (which corresponds to the $\ell_1$ norm) implies solving the corresponding problem for any $\ell_p$ norm, for $p \in [1,\infty]$, as $\|x-y\|_p \leq \|x-y\|_1$. Another reason is that for many other metrics, the sample complexity of a hypothesis selection problem can depend on structural properties of the reference class $\Q$, which is undesirable for formulating problem-independent theorems like \autoref{thm:main}. For a more elaborate discussion of the advantages in working with total variation, see Chapter~$6.5$ in~\cite{Devroye01combinatorial}, and Section~$3.1$ in \cite{Ashtiani18mixtures}.

We believe that our technique can be extended to derive hypothesis selection algorithms for other distance measures that satisfy (at least some approximate) version of the triangle inequality\footnote{See \Cref{sec:bkm} for our usage of the triangle inequality.} ({\em e.g.}, Hellinger distance and other metric spaces).

\paragraph{Proper {\em vs.}\ improper.}

A basic classification of machine learning problems distinguishes between {\it proper} and {\it improper}
learning. In the proper case the algorithm always outputs a distribution $q \in \Q$, whereas in the improper
case it may output an arbitrary distribution. Improperness has been shown to be beneficial in many settings (see, {\em e.g.},\,\cite{schapire2012boosting,Daniely14proper}), including the agnostic hypothesis selection setting: while Yatracos's $3$-approximation algorithm is proper, \cite{BKM19} prove that the factor~$3$ cannot be improved by any proper algorithm (with any sample complexity)\footnote{We mention that for the case $n=2$, a proper $2$-approximation algorithm for the agnostic hypothesis selection problem was given by \cite{Mahalanabis08density}.}. For this reason, their and our $2$-approximation algorithms are inherently improper. For many applications ({\em e.g.}, applications to density estimation discussed above), improper hypothesis selection algorithms suffice.

\paragraph{{Computational complexity.}}

Although our approach is algorithmic, our focus is not on computational efficiency.  
While the sample complexity of our algorithm is only logarithmic in the number of distributions $n$ (and is independent of the domain size $|\X|$), in the general case, its running time scales polynomially with both $n$ and $|\X|$, as is the case for other sample-efficient hypothesis selection algorithms. Clearly, the dependence on $n$ cannot be sub-linear (each $q_i$ needs to be accessed, unless some structure on $\Q$ is assumed). As for the dependence on $|\X|$, our algorithm assumes oracle access to operations on~$\X$, such as checking membership in sets of the form $F = \{ x \in \cX: \; q_1(x) > q_2(x) \}$\footnote{These are the, so called, ``Yatracos sets'' and Yatracos's algorithm also assumes membership oracle to them.}, and several other (somewhat involved) operations\footnote{In the language of the overview presented in \Cref{sec:overview}, these operations include finding a distribution $q$ such that $v(q) \leq v$, and solving the optimization problem corresponding to finding the discriminating sets~$F_i$.} that can only be implemented efficiently for restricted classes~$\Q$. We mention that the situation is similar for many density estimation problems: the existence of polynomial time algorithms is unknown even for specific natural classes, such as mixtures of gaussians (see \cite{Ashtiani18mixtures} for further discussion).

While efficient algorithms ({\em e.g.}, with $\poly\log(|\X|)$ running-time) for all classes $\Q$ are unlikely in the simple and abstract learning setting considered by this work, this setting is particularly suited to capture basic information-theoretic resources, such as sample-complexity and approximation guarantees, which are not affected by the computational model. As discussed above, the complexity of these resources is still poorly understood, even for very basic problems.


\subsection{Additional Related Work} \label{sec:related}

In this work we give a novel approximation algorithm for {hypothesis selection} of any (finite) class~$\Q$, following the classical work of \cite{yatracos85, devroye1996universally, devroye1997nonasymptotic, Devroye01combinatorial} and the recent work of \cite{BKM19}, discussed above. Over the last decade or so, hypothesis selection received quite a bit of attention by different theoretical communities  and many aspects of this problem were studied, including computational efficiency, robustness, weaker access to hypotheses, privacy and more (see, {\em e.g.}, \cite{Mahalanabis08density, DDS15, DK14, SOAJ14, AJOS14, Chan14histograms, DKKL19, BKSW21, AFJOS18, BKSW21, GKNWZ20}).

Hypothesis selection can also be viewed as a special case of density estimation (also known as distribution learning), where one wishes to learn a (typically infinite) class of densities from samples. In fact, as mentioned above, many density estimation algorithms use hypothesis selection algorithms as fundamental subroutines. Density estimation is a very basic unsupervised learning problem studied since the late nineteenth century, starting with the pioneering work of Pearson~\cite{Pearson95contributions}. Since, it was systematically studied for many natural classes, such as mixtures of gaussians ({\em e.g.},~\cite{Kalai12mixtures, Diakonikolas17mixtureslower, Diakonikolas18mixtures, Kothari18robust, Ashtiani18mixtures1, Ashtiani18mixtures}), 
histograms ({\em e.g.}, ~\cite{Pearson95contributions, lugosi96histograms, Devroye04bin, Chan14histograms, Diakonikolas18histograms}), and more. For a fairly recent survey see~\cite{Diakonikolas16survey}.


Our result yields improved approximation guarantees in many of these works. For example, plugging it in \cite{Ashtiani18mixtures}, instead of Yatracos's algorithm which is used as a black box, improves the approximation factor from $3$ to $2$ for learning gaussians, and from $9$ to $6$ for learning mixtures of gaussians, while keeping the sample complexity near-optimal.

\paragraph{Optimization and online learning.}
{
A key component in our derivation is the cutting-with-margin game.
	This game is reminiscent of dynamical processes which are studied in optimization and online learning.
	In particular, our solution to this game is based on a greedy approach of maximizing	the entropy and a potential-based analysis 
	which brings to mind standard $\kl$-divergence-based analyses of mirror-decent and multiplicative-weights update (see, {\em e.g.},~\cite{Azoury01relative,Arora12mw,Bubeck15convex}).
{Moreover, the cutting-with-margin game naturally generalizes to arbitrary norms $\|\cdot \|$ by replacing
	the $\ell_1$ norm with $\|\cdot\|$ and the simplex $\Delta_n$ by the unit ball with respect to $\|\cdot\|$.
	One can extend our upper bound to arbitrary norms, 
	by replacing the $\kl$-divergence with an appropriate {\em Bregman divergence}\footnote{ Using the Bregman divergence, we have some preliminary results regarding the round complexity of our cutting-with-margin game in other norms. These include a nearly tight bounds for the $\ell_p$ norm, when $p \in (1,2]\cup\{\infty\}$:
if $p\in(1,2)$ then the player can solve the corresponding game in $r = O_p(1/\eps^2)$ rounds, and if $p=\infty$  a
then the round complexity of the game is $\Theta(n\log(1/\eps))$.}, as is the case for some optimization problems.}
	
These technical interrelations suggest the possibility of a deeper connection between
	the cutting-with-margin game and online optimization.
	Ideally, one could hope to find a formal reduction by phrasing
	our game as a convex regret minimization problem.
	We remark, however, that, unlike regret minimization problems, our game is not defined via a local regret function, but rather defined using a very global cost function. 
	We leave this further exploration of the relations between our game to the regret minimization framework for future work.
	
\paragraph{The ellipsoid method.} Another known algorithm that is of a particular syntactic similarity to our cutting-with-margin game is the well-known {\em ellipsoid method} for solving {\em linear programs}: in both settings a player maintains a convex set in $\mathbb{R}^n$ (in our game it is, without loss of generality, a polytope, and when running the ellipsoid method it is an ellipsoid), and in each step it selects a point within that set. If the selected point is not a ``solution'', the player receives a separating hyperplane from an adversary or a hyperplane oracle, which separates the selected point from the target set of solutions. Then, the player moves to a ``smaller'' convex body that lies, in its entirety, on one side of the hyperplane.

We note that a crucial difference between the two is that when running the ellipsoid method, the ellipsoids are getting rapidly smaller in terms of {\em volume} (and, for example, the next ellipsoids need not be contained in the former one), and it is this decrease in volume that allows for a fast convergence. In contrast, as will be discussed in \Cref{sec:cuttingsolution}, shrinking the volume of our convex body between rounds of the cutting-with-margin game does not suffice for convergence (and therefore, ``{\em  centroid}-based'' methods do not apply).

\section{Proof Overview} \label{sec:overview}

In this section we overview the proofs and highlight some of the more technical arguments. We defer the full proof to the Appendix.

Let $\cQ=\{q_1,\ldots, q_n\}$ be a (known) finite reference class of distributions and let $p$ denote the target distribution to which we have sample access. Denote $i^\star = \arg\min_i\{\tv(p,q_i)\}$. Our goal is to use as few samples as possible from~$p$ in order to find $q$ such that $\tv(p,q) \leq 2 \cdot \tv(p,q_{i^\star}) + \eps$.

\subsection{A Geometric Approach to Hypothesis Selection}\label{sec:bkm}

Our starting point is the $2$-approximation algorithm of \cite{BKM19}. In this subsection we describe our interpretation of their technique (some of the claims we make here are implicit in their paper).

The basic observation of \cite{BKM19} is that it suffices to find a distribution $q$ which is (almost) at least as close to each of the $q_i$'s as $p$,
\begin{equation} \label{eq:sketch-q}
(\forall i): \tv(q,q_i)\leq \tv(p,q_i) + \eps.
\end{equation}
Finding such a $q$ suffices, as by the triangle inequality, $\tv(q,p)\leq \tv(q,q_i) + \tv(q_i,p) \leq 2\tv(q_i,p) + \eps$ for every $i$, and, in particular, for $i^\star$. 

This suggests the following definitions: for a distribution $q$, let $v(q)\in[0,1]^n$ denote the vector of all distances $v(q)=(\tv(q,q_i))_{i=1}^n$; a vector $v\in [0,1]^n$ is {\it feasible} if $v\geq v(q)$ for some distribution~$q$ (when we write $u \geq w$ for $u,w \in [0,1]^n$ we mean $(\forall i):\; u_i \geq w_i$). With this notation, our goal is to find $v$ such that 
\begin{itemize}
\item[(i)] $v \leq v(p) + \eps\cdot 1_n$, where $1_n$ is the all-one vector, and 
\item[(ii)] $v$ is feasible. 
\end{itemize}
Once such a vector $v$ is obtained, one can find a distribution $q$ satisfying~$v(q)\leq v$, and consequently a $2$-approximation for the target distribution $p$.
        
Let $\cP\subseteq [0,1]^n$ denote the set of all feasible vectors $v$ and note that it is convex and upward-closed.  
The approach of \cite{BKM19} for finding a desired $v$ proceeds in rounds, where in round $k$ we find a vector $u_k$ that is closer to the feasible set, while maintaining the invariant that $u_k \leq v(p)$:
 
\begin{enumerate}
	\item Let $u_0= \vec 0\in [0,1]^n$ be the all-zero vector. Note that $u_0\leq v(p)$, so $u_0$ satisfies the above Item~(i), but not Item (ii) (except in trivial cases).
	\item For $k=0,1,\ldots$
\begin{enumerate}
	\item If $u_k + \eps\cdot 1_n$ is feasible (that is, if $d_\infty(u_k,\cP) \leq \eps$, where $d_\infty(\cdot,\cdot)$ denotes $\ell_\infty$ distance), then output a $q$ such that $v(q) \leq u_k + \eps \cdot 1_n$ ($\leq v(p) + \eps \cdot 1_n$).
	\item \label{item:2b} Else, use samples from $p$ to derive $u_{k+1}$ such that $u_k \leq u_{k+1} \leq v(p)$, and $u_{k+1}$ is ``closer'' (in some measure, see below) to $\cP$.
\end{enumerate}
\end{enumerate}

\paragraph{Selecting the new point $u_{k+1}$.} The crux of this approach is the update step in which $u_{k+1}$ is computed given $u_k$. Since $d_\infty(u_k,\cP) > \eps$, there exists a $u_{k+1}$ such that $u_k \leq u_{k+1} \leq v(p)$ and $d_1(u_{k+1},u_{k}) \geq \frac{\eps}{2}$ (for instance, since there exists a coordinate $i \in [n]$ such that $u_k + \frac{\eps}{2} \cdot e_i < v(p)$, where $e_i$ is the $i^{\text{th}}$ unit vector). \cite{BKM19} show how to find such a $u_{k+1}$ with few queries (discussed next), and they use this $u_{k+1}$ as their next point. However, since $\| 1_n \|_1 = n$, their strategy may require $\Omega(\frac{n}{\eps})$ rounds.

\subsubsection{Implementing the Strategy}\label{sec:minimax}

\paragraph{Violated tests.} We next explain how \cite{BKM19} find the coordinate $i$ of $u_k$ that they wish to update. To this end, observe that whenever $u_k + \eps \cdot 1_n$ is not feasible there is a hyperplane separating the point $u_k+\eps\cdot 1_n$ from the set~$\cP$ of feasible vectors, witnessing the fact that  $d_\infty(u,\cP)>\eps$. We call a normal $h\in \Delta_n$ to such a hyperplane a ``{\it violated test}'' 
(here $\Delta_n$ denotes the simplex of all probability vectors in $\R^n$). 
For $u \in [0,1]^n$ and $d >0$, we denote the set of all violated tests witnessing the fact that $u + d \cdot 1_n$ is not feasible by 
\[\cH_d(u) = \Bigl\{h\in \Delta_n : \;  h \cdot u + d  < \min_{v\in \cP} h \cdot v \Bigr\}.\]
	
\paragraph{From a test $h$ to an updated point $u_{k+1}$.} We next informally state a central lemma proved by \cite{BKM19}, showing how to convert any violated test $h$ to a new point $u_{k+1}$ (for a precise statement, see Lemma 12 in \cite{BKM19} or \Cref{lem:progress} in this paper).

\begin{lemma}\label{eq:sketch-lemma}
Using $n$ statistical queries (queries of the form $p(F)$ for some set $F$), any $h \in \cH_\eps(u_k)$ can be converted to a point $u_{k+1}$ satisfying:
\begin{enumerate}
\item \label{eq:sketch-smaller-than-v-p} $u_k \leq u_{k+1} \leq v(p)$.
\item \label{eq:sketch-H-eps-2} $u_{k+1}$ passes the test induced by $h$: $h \notin \cH_{\frac{\eps}{2}}(u_{k+1})$. This also implies that $h \cdot (u_{k+1} - u_k) > \tfrac{\eps}{2}$ (as $h \in \cH_{\eps}(u_k)$ implies $h \cdot u_k + \eps  < \min_{v \in \cP} h \cdot v$ and $h \notin \cH_{\frac{\eps}{2}}(u_{k+1})$ implies $h \cdot u_{k+1} + \frac{\eps}{2}  \geq \min_{v \in \cP} h \cdot v$). 
\end{enumerate}
\end{lemma}

Observe that the $u_{k+1}$ constructed by this lemma (for any $h$) satisfies $d_1(u_{k+1},u_{k}) \geq \frac{\eps}{2}$ (due to \Cref{eq:sketch-H-eps-2}, recall that $h \in \Delta_n$), and therefore it can be used to implement the strategy of \cite{BKM19}.

\paragraph{Proving the lemma.} While the proof of \Cref{eq:sketch-lemma} is pretty short, it is tricky. For completeness, we will next give some intuition for it by showing how to construct $u_{k+1}$ for a specific (easy to handle)~$h$. 

Assume that $u_k+\eps \cdot 1_n$ is not feasible and that $h = (\frac{1}{2},\frac{1}{2},0,\ldots,0) \in \cH_\eps(u_k)$. Denote $F=F(q_1,q_2) = \{x : q_1(x) \geq q_2(x)\}$. (Observe that this is the so-called {\it Yatracos set} which is used in Yatracos's $3$-approximation algorithm and satisfies $\tv(q_1,q_2) = q_1(F)- q_2(F)$). Use samples from $p$ to get an estimate $\hat p(F)$ of $p(F)$ up to an $\frac{\eps}{4}$ additive term. Set $z_i = \lvert \hat p(F) - q_i(F) \rvert - \frac{\eps}{2}$ for $i=1,2$ and $z_i=0$ for $i \geq 3$. Obtain $u_{k+1}$ from $u_k$ by setting $(u_{k+1})_i = \max\{(u_k)_i,z_i\}$.
        
The resulting $u_{k+1}$ satisfies \Cref{eq:sketch-smaller-than-v-p}, as since $\lvert p(F) - q_i(F)\rvert \leq \tv(p,q_i) = (v(p))_i$ it follows that $z_i \leq (v(p))_i$. 
It also satisfies \Cref{eq:sketch-H-eps-2}, as 
\begin{align} \label{eq:sketch-u_k_plus_1}
h \cdot u_{k+1} + \tfrac{\eps}{2} 
&= \tfrac{1}{2} ((u_{k+1})_1 + (u_{k+1})_2) + \tfrac{\eps}{2} \geq \tfrac{1}{2} (z_1 + z_2) + \tfrac{\eps}{2} \\
&\geq \tfrac{1}{2}(\lvert \hat p(F) - q_1(F) \rvert + \lvert \hat p(F) - q_2(F) \rvert) 
\geq \tfrac{1}{2} \lvert q_1(F) - q_2(F) \rvert = \tfrac{1}{2}\tv(q_1,q_2)
= \min_{v\in \cP} h \cdot v, \nonumber
\end{align}
where the last equality is because for every $v = v(q) \in \cP$ it holds that $h \cdot v = \frac{1}{2} (v_1 + v_2) = \frac{1}{2} (\tv(q,q_1)+\tv(q,q_2)) \geq \frac{1}{2} \tv(q_1,q_2)$ and for $v = v(q_1) \in \cP$ it holds that $h \cdot v =  \frac{1}{2} \tv(q_1,q_2)$. 

\paragraph{Query/sample complexity.}

For a general $h$, the proof of the lemma is more involved and crucially relays on the Minmax theorem. The point $u_{k+1}$ is computed as $(u_{k+1})_i =  \max\{(u_k)_i,z_i\}$, where for every $i \in [n]$, $z_i$ is of the form $z_i = \lvert \hat p(F_i) - q_i(F_i) \rvert - \frac{\eps}{2}$, for some set $F_i$ and where $\hat p(F_i)$ is an approximation of $p(F_i)$ to within an additive error of $c \cdot \eps$ for some constant $c < 1$. 

Computing $u_{k+1}$ requires $n$ statistical queries (the values of $p(F_i)$ for all $i$'s), where each needs to be approximated to within an additive error of $c \cdot \eps$. While approximating each query separately requires $\Theta(1 /\eps^2)$ samples, by a standard combination of Chernoff and union bound, all $n$ queries can be approximated using $O(\log n /\eps^2)$ samples.

\subsection{The Cutting-With-Margin Game:~A~Dual~Perspective}\label{sec:cutting}

Recall that we wish to find a rule for updating $u_k$ to a $u_{k+1}$ satisfying $u_k < u_{k+1} < v(p)$ that will allow us to reach a feasible point after the minimum number of steps. We wish to define a measure of progress to help us choose our next $u_{k+1}$. As discussed above, \cite{BKM19} use the $\ell_1$ norm as their measure of progress, but this results in a slow convergence to a feasible point. 

To find a better progress measure, we revisit \Cref{eq:sketch-lemma}, specifically \Cref{eq:sketch-H-eps-2} that shows that by updating $u_k$ using the test $h \in \cH_\eps(u_{k})$, it is not only that $h \notin \cH_\eps(u_{k+1})$, but also $h \notin \cH_{\frac{\eps}{2}}(u_{k+1})$. We interpret this as implying that the set of violated tests can shrink substantially between rounds. This suggests a new approach: instead of measuring progress by comparing the locations of $u_k$ and~$u_{k+1}$, we can take a ``{\em dual}'' view and compare the sizes of the sets $\cH_{\eps}(u_k)$ and $\cH_{\eps}(u_{k+1})$ of violated tests that we still need to rule out (recall that if this set is empty, we have found a feasible point). We note that this ``dual'' view is lossy (and is not a dual in the standard sense) as the mapping $u_k \to \cH_{\eps}(u_k)$ may not be one-to-one.

\paragraph{The cutting-with-margin game.} Consider a sequence $\vec 0 = u_0 \leq u_1 \leq \ldots \leq u_m$ in which the point $u_{k+1}$ was produced from $u_k$ by selecting some $h_k \in \cH_{\eps}(u_k)$ and applying \Cref{eq:sketch-lemma}, and where $u_m$ is feasible. Denote $\cH_k = \cH_{\eps}(u_k)$. It can be shown that $\cH_k$ is convex for every $k$, and that $\cH_{0} \supset \cH_{1} \supset \cH_2 \supset \ldots \supset \cH_m = \emptyset$ ($\cH_m = \emptyset$ as $u_m$ is feasible). Furthermore, we are able to prove that $\cH_{k+1}$ is disjoint from an $\ell_1$ ball of radius $\Omega(\eps)$ around~$h_k$ (see \Cref{lemma:H-cap-B-empty}). Intuitively, this is because $h_k \notin \cH_{\frac{\eps}{2}}(u_{k+1})$ (\Cref{eq:sketch-lemma}, \Cref{eq:sketch-H-eps-2}) implies that the generated $u_{k+1}$ not only passes the test induced by $h_k$, but also passes all ``similar'' tests.

The above discussion gives rise to the {\it cutting-with-margin} game discussed in the introduction (see \Cref{sec:intro-game}).
Recall that this is a game between a { player} and an {adversary}, and it is played over a convex body~$\cH \subseteq \Delta_n$ known to both the player and the adversary. 
Let $\cH_0= \cH$;  in every round $k=0,1,\ldots$ of the game, the player selects a point $h_k \in \cH_k$ and the adversary picks $\cH_{k+1} \subseteq \cH_k$
to be any convex set which is disjoint from the $\ell_1$ ball of radius $\eps$ around $h_k$. The game ends when the set $\cH_k$ is empty. See illustration in \Cref{fig:cutting-with-margin}. Of course, the task is now to find a strategy that solves this game with minimum number of rounds. Note that, in the language of this game, the strategy of \cite{BKM19} selects an arbitrary $h_k \in \cH_{\eps}(u_k)$ in round $k$. We will next show a strategy for selecting $h_k$ that will allow for a faster convergence.

\subsection{Warm-up: $\poly(\log n /\eps^2)$ Sample Complexity}\label{sec:cuttingsolution}

So far, we reduced the hypothesis selection problem to solving the cutting-with-margin game. We next outline a solution for the cutting-with-margin game in $\tilde O(\log n /\eps^2)$ rounds. Since the implementation of each round requires $O(\log n /\eps^2)$ samples (see \Cref{sec:minimax}), this implies an algorithm for hypothesis selection with $\tilde O(\log ^2 n/\eps^4)$ sample complexity.
 
First observe that an equivalent way of presenting the cutting-with-margin game lets the adversary pick in each round a halfspace $H_k$ which is disjoint from the $\ell_1$ ball of radius $\eps$ around~$h_k$, and the game continues with $\cH_{k+1}= \cH_k\cap H_k$. This presentation is reminiscent of {\em Grunbaum's inequality}~\cite{grunbaum60partitions}, which guarantees that if the player picks the {\em centroid} (which is a standard way of defining the ``center'' of a body) of $\cH_k$ then $vol(\cH_{k+1}) \leq (1-e^{-1})\cdot vol(\cH_k)$, 	where $vol(\cdot)$ is the standard (Lebesgue) volume. While the centroid is an intuitive choice for our player, a counter strategy by the adversary will pick bodies that have small volumes but large diameters. Indeed, note that as long as the diameter of the body is greater than $\eps$, the adversary can force at least one additional round. This shows that the volume is too crude of a measure for our game. Ideally, we would have wanted to use a different ``centroid'' that satisfies an analogous property with respect to the diameter (say, $diameter(\cH_{k+1}) \leq \frac{99}{100}\cdot diameter(\cH_k)$). Unfortunately, no such object exists. 

The approach we take for designing our player stems from the observation that if the player could always pick a point $h_k\in \Delta_n$ that is close to the uniform distribution $h^\star=(\frac{1}{n},\ldots,\frac{1}{n})$, then the game would have been solved in a few rounds. It is the easiest to see why when using the ``primal'' point of view from \Cref{sec:bkm}: indeed, assume $u_k + \eps\cdot 1_n$ is separated from $\cP$ by a hyperplane perpendicular to $h^\star=(\frac{1}{n},\ldots,\frac{1}{n})$. Then, since $u_{k+1}\geq u_k$ lies on the other side of that hyperplane, it follows that $\lvert u_{k+1}- u_k\rvert_1\geq \eps n$. So, when updating from $u_k$ to $u_{k+1}$, the~$\ell_1$ norm increases by at least~$\eps n$ (recall from \Cref{sec:bkm} that in the \cite{BKM19} strategy the $\ell_1$ norm increases by only $\Omega(\eps)$ in each round). Thus, since in $[0,1]^n$ the $\ell_1$ norm is bounded by $n$, the total number of such steps is at most $O(1/\eps)$. Of course, this strategy is impossible, as if $h_1 = h^\star$ then a ball of radius $\eps$ is disjoint from $\cH_{k}$, for all $k>1$.

\paragraph{Entropy as a progress measure.} Inspired by the above intuition, our approach will be to set $h_k \in \cH_k$ to be as ``close'' to $h^\star$ as possible. Indeed, we select $h_k \in \cH_k$ that maximizes the {\em entropy} function (here we view the point $h_k \in \Delta_n$ as a distribution). This corresponds to measuring the distance from the uniform distribution $h^\star$ using $\kl$-divergence.  The reason that the entropy function gives an efficient solution for our game boils down to that it is 
	(i) {\em strongly convex} w.r.t $\ell_1$ (as is evident by {\it Pinsker's Inequality}), 
	(ii) {\em bounded} by $\log(n)$ over the simplex.  
Roughly speaking, strong convexity means that in every step the entropy drops by~$\Omega(\eps^2)$. 
This, combined with the fact that the entropy is bounded by $\log(n)$, implies our $\tilde O(\log(n)/\eps^2)$ solution for the cutting-with-margin game\footnote{Given that, it is natural to look for a strongly convex function over the simplex that is bounded by $\ll \log(n)$. However, no such function exists.}.
	
As discussed in the introduction, entropy and $\kl$-divergence based strategies are often used in the context of optimization and regret minimization, basically for similar reasons (convexity and boundedness). However, our game is not defined by a cost function measuring the cost of each round separately, but rather, our ``cost function'' is the length of the game.  

\begin{figure}
\begin{center}
\includegraphics[width=.6\textwidth]{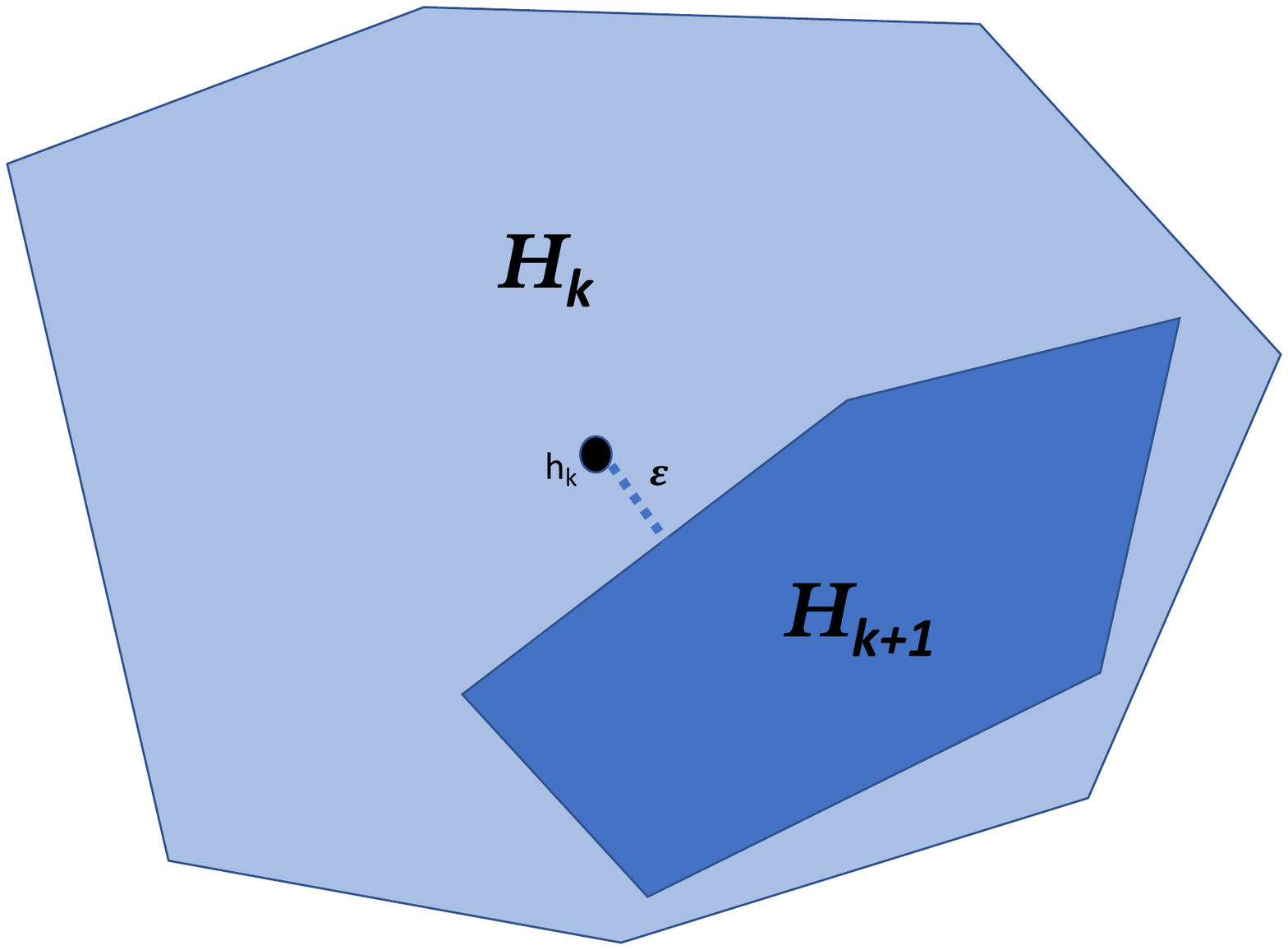}
\end{center}
\caption{\small An illustration of the cutting-with-margin game:
in each step $k$ the player picks a point $h\in \cH_k$ and announces it to the adversary.
The adversary then replies with $\cH_{k+1}\subseteq \cH_k$ which is convex and disjoint from an $\ell_1$ ball of radius $\eps$ around $h_k$. The players' goal is to empty the set as fast as possible ({\em i.e.},\ to reach $\cH_k=\emptyset$), and the adversary's goal is to delay the player.}
\label{fig:cutting-with-margin}
\end{figure}

\subsection{Near-Optimal Sample Complexity}\label{sec:optoverview}

In \Cref{sec:cuttingsolution}, we gave a hypothesis selection algorithm with $\tilde O(\log^2  n /\eps^4)$ samples, by solving the dual game. While this algorithm uses exponentially less samples than the one by \cite{BKM19}, it still sub-optimal. We next show how to obtain an algorithm with a near-optimal sample complexity of $\tilde O(\log  n /\eps^2)$, by first improving the dependence on $n$ to $\tilde O(\log n)$ (less involved), and then improving the dependence on $\eps$ to $O(1/\eps^2)$ (one of the main technical contributions of this paper). Since the sample complexity of our resulting algorithm (almost) matches Yatracos's, it can replace Yatracos's algorithm in density estimation algorithms to obtain a better approximation factor, while keeping the same low sample complexity.

\subsubsection{Optimal Dependence on $n$} 

We revisit the basic observation from \Cref{sec:bkm} that finding a distribution $q$ satisfying $(\forall i): \tv(q,q_i)\leq \tv(p,q_i) + \eps$ suffices in order to get a $2$-approximation for hypothesis selection (see \Cref{eq:sketch-q}). We observe that it also suffices to find $q$ that only satisfies $\tv(q,q_{i^\star}) \leq \tv(p,q_{i^\star}) + \eps$ (recall that $i^\star$ minimizes $\tv(p,q_{i})$) for exactly the same reason: $\tv(q,p)\leq \tv(q,q_{i^\star}) + \tv(q_{i^\star},p) \leq 2\tv(q_{i^\star},p) + \eps$. Thus, it suffices for our algorithm to maintain the invariant $(u_k)_{i^\star} \leq (v(p))_{i^\star}$, instead of $u_k \leq v(p)$. This suggests that we can relax \Cref{eq:sketch-smaller-than-v-p} in \Cref{eq:sketch-lemma} and only require $(u_{k+1})_{i^\star} \leq (v(p))_{i^\star}$ (in addition to $u_k \leq u_{k+1}$). 

Due to the above, had we known $i^\star$, we would only shoot for a good approximation (to within~$c \cdot \eps$) of $(u_{k+1})_{i^\star}$, which means that \Cref{eq:sketch-lemma} can use only $O(1/\eps^2)$ samples (to get a good approximation of $p(F_{i^\star})$). But, we {\em don't know the identity of $i^\star$}. The crucial observation here is that this does not matter. We can use {\em the same} $O(1/\eps^2)$ samples to evaluate each of the~$n$ statistical queries corresponding to each of the coordinates of $u_{k+1}$. Of course, since we are using too few samples, some of these coordinates will not be well approximated. However, it is likely that each one by itself will, and, in particular, this will be the case for $(u_{k+1})_{i^\star}$. In other words, since we only care about $(u_{k+1})_{i^\star}$, we no longer have to pay for a costly union bound over all $n$ coordinates. (We also  show that \Cref{eq:sketch-H-eps-2} in \Cref{eq:sketch-lemma} still holds under this approximation using an averaging argument).

\subsubsection{Optimal Dependence on $\eps$} 

Recall that in each step of the cutting-with-margin game, the player picks a point $h_k \in \cH_k$, and the adversary sets $\cH_{k+1}\subseteq \cH_k$ by cutting away an $\ell_1$ ball of radius $\eps$ around $h_k$. The algorithm we have so far uses $\Omega(\log n/\eps^4)$ samples from $p$: every round uses $\Theta(1/\eps^2)$ samples and $\max_{h \in \cH_\eps(u_k)} \{ \bH(h) \}$ drops by $\Omega(\eps^2)$ (recall that, to begin with, the entropy is at most $\log n$ and we want it to drop to~$0$). 

To reduce the sample complexity, we move away from this ``static'' type of algorithms and design a  ``dynamic'' algorithm whose number of samples per round may vary (but, will never exceed $\Omega(1/\eps^2)$). The important property of the new algorithm is that {\it if the algorithm samples more points from $p$, then the adversary cuts away a larger $\ell_1$ ball around $h_k$.} Specifically, if~$O(1)$ points are sampled then the radius of the removed ball is $\eps$, and if $O(1/\eps^2)$ points are samples then the radius removed ball will be $\Omega(1)$. We will show that this coupling of the number of samples used in a step with the amount of progress made in that step (instead of using the maximum number of samples in every step and expecting the minimum progress) enables a {\em win-win} analysis which implies the desired saving in the sample complexity.

\paragraph{Bounding the radius of the removed ball.} 

To explain how this idea is implemented, we need to dive into the details of the algorithm. Recall that the algorithm aims to find a point $v$ such that $v_{i^\star} \leq \tv(p,q_{i^\star}) + \eps$, and for which $\cH_\eps(v) = \emptyset$. Assume that the current point $u_k$ satisfies $d_\infty(u_k,\cP)=d \gg \eps$ (which means $\cH_{d}(u_k) = \emptyset$) and that we aim at reducing the distance to, say,~$\frac{3d}{4}$. That is, we want to get to a point $u$ such that $d_\infty(u,\cP) \leq \frac{3d}{4}$, or, equivalently, $\cH_{\frac{3d}{4}}(u) = \emptyset$. Recall from \Cref{sec:bkm} that towards this, we pick a violated test $h_k \in \cH_{\frac{3d}{4}}(u_k)$ which, by applying \Cref{eq:sketch-lemma}, yields the new point $u_{k+1} \in [0,1]^n$. Of course, the lemma uses samples from $p$ to compute this $u_{k+1}$. As we soon see, in some cases it will be worthwhile for our algorithm to only compute a crude approximation of this $u_{k+1}$ using fewer samples. Part of the difficulty is to decide on the quality of this approximation without knowing~$u_{k+1}$.

Nevertheless, imagine for a moment that the algorithm does know this~$u_{k+1}$ and uses it as its next point. How much ``progress'' does this imply in the cutting-with-margin game? That is, how much smaller is $\cH_{\frac{3d}{4}}(u_{k+1})$ compared to $\cH_{\frac{3d}{4}}(u_k)$? Denote $w_k = u_{k+1} - u_k$. We next show that $\cH_{\frac{3d}{4}}(u_{k+1})$ is disjoint from an $\ell_1$ ball of radius 
\begin{equation} \label{eq:sketch-r}
r = \frac{d}{8 \|w_k\|_\infty}
\end{equation} around $h_k$ (we wish for $r$ to be as large as possible).
Intuitively, if $\|w_k\|_\infty$ is small, it means that we have made progress in many coordinates (though the progress in each might be relatively small). 
Since we are getting close to $\cP$ in many directions, this should imply that $u_{k+1}$ passes many of the tests $h_k$ that were violated by $u_k$, and thus that $\cH_{\frac{3d}{4}}(u_{k+1})$ is much smaller.

More formally, let $h\in \cH_{\frac{3d}{4}}(u_{k+1})$, \Cref{eq:sketch-r} follows from:
\begin{align*}
\|h_k-h\|_1 \cdot \|w_k\|_\infty \geq (h_k-h)\cdot(u_{k+1}-u_k) \geq \tfrac{d}{8}.
\end{align*}
Here, the first inequality is due H\"older's Inequality. The second inequality is because $h_k \cdot (u_{k+1} - u_k) \geq \frac{3d}{8}$ (due to \Cref{eq:sketch-lemma}, \Cref{eq:sketch-H-eps-2}) and because $h\cdot (u_{k+1}-u_k) \leq \frac{d}{4}$ (since $h\in \cH_{\frac{3d}{4}}(u_{k+1})$ it holds that $h\cdot u_{k+1} + \frac{3d}{4} < \min_{v\in \cP}h\cdot v$, while since $h \notin \cH_{d}(u_{k}) = \emptyset$ it holds that $h \cdot u_{k} +  d \geq \min_{v\in \cP}h\cdot v$). 

\paragraph{Our ``win-win'' strategy.} The take home message from the above discussion is that:
\begin{center}
{\it If $\|w_k\|_\infty$ is small then $\cH_{\frac{3d}{4}}(u_{k+1})$ is small.}
\end{center}
We next show that this relation leads us to a ``{\em win-win}'' situation: if $\|w_k\|_\infty$ is large, it suffices to only crudely approximate $w_k$, and we save on samples. However, if $\|w_k\|_\infty$ is small, $\cH_{\frac{3d}{4}}(u_{k+1})$ is small and we made a lot of progress towards ruling out all violated tests. 

To see the relation between $\|w_k\|_\infty$ and the number of samples required to approximate $w_k$, first assume that $w_k$ is uniform over a set of coordinates of size $m$ ({\em i.e.}, for every $i \in [n]$, either $(w_k)_i = 1/m$ or $(w_k)_i = 0$). Now, if $m$ is small than all non-zeros coordinates of $w_k$ are large, and thus~$w_k$ can be reasonably approximated with few samples. (In fact, the number of samples scales with~$(1/\|w_k\|_\infty)^{2}$).
	
\paragraph{Slicing.} Of course, $w_k$ may not be uniform on a set. To deal with such $w_k$'s, we partition $w_k$ to $\log(1/d)$ many ``slices'' $w_k = w^1_k+\ldots+w^{\log(1/d)}_k$ such that each $w^\ell_k$ is almost uniform over a set (specifically, for $\ell < \log(1/d)$, each of the coordinates of $w^\ell_k$ is either $0$ or in $(2^{-\ell},2^{-(\ell-1)}]$). 
We then try to identify a slice with a significant contribution to $h_k \cdot w_k = \sum_{\ell \in [\log(1/d)]} h_k \cdot w^\ell_k$ (recall that $h_k \cdot w_k \geq \frac{3d}{8}$ due to \Cref{eq:sketch-lemma}, \Cref{eq:sketch-H-eps-2}).
However, since $w_k$ is not known to the algorithm, we use samples to learn it ``slice-by-slice'', starting by approximating $w^1_k$, the slice containing the largest values and requiring the least number of samples to estimate, and continuing to the slices that require more samples, until reaching a ``good'' slice. We mention that this slice-searching process is equivalent to playing the dual game with different $\eps$ values.

\section{Preliminaries}

\subsection{Notation}
Let $n \in \N$. For $u,v \in \R^n$, we write $u \geq v$ if $(\forall i \in [n]): \; u_i \geq v_i$. We use $u \cdot v := \sum_{i \in [n]} u_i v_i$ to denote the standard inner product of $u$ and $v$. 

For $p\in [1,\infty]$,  we denote by $\| \cdot \|_p$ the $\ell_p$ norm.
For~$u\in \R^n$ and $r \geq 0$, let $B_p(u,r)$ denote a ball of radius $r$ with respect to $\ell_p$ that is centered at~$u$,
 $$B_p(u,\eps) = \{v \in \R^n :\; \|v - u\|_p \leq r \}.$$


Let $\Delta_n$ denote the simplex of probability vectors in $\R^n$, 
$$\Delta_n := \bigl\{h\in \R^n : \sum_{i \in [n]} h_i = 1, ~ (\forall i): h_i \geq 0\bigr\}.$$

The entropy function is denoted by $\bH$ and the Kullback-Leibler divergence by $\kl$.

\subsection{Definition of the Hypothesis Selection Problem}
Let $\X$ be a domain and let $\Delta(\X)$ denote the set of all probability distributions over $\X$.
We assume that either (i) $\X$ is finite in which case $\Delta(\X)$
is identified with the set of $\lvert \X\rvert$-dimensional probability vectors, or
(ii) $\X=\R^d$ in which case $\Delta(\X)$ is the set of Borel probability measures.


Let $\Q\subseteq \Delta(\X)$ be a set of distributions.
We focus on the case where $\Q$ is finite and denote its size by~$n$.
Let~$\alpha > 0$, we say that $\Q$ is {\it $\alpha$-learnable with sample complexity} $m(n,\eps,\delta)$ if there is a {(possibly randomized)} algorithm $A$ such that
for every $\eps,\delta>0$ and every target distribution $p\in \Delta(\X)$,
if $A$ receives as input at least $m(n,\eps,\delta)$ independent samples from $p$
then it outputs a distribution $q$ such that
\[\tv(p,q)\leq \alpha\cdot\opt + \eps,\]
with probability at least $1-\delta$, where $\opt = \min_{q\in \Q}\tv(p,q)$
and $\tv(p,q) = \sup_{A\subseteq \X}\{p(A)-q(A)\}$ is the total variation distance.
We say that $\Q$ is {\it properly $\alpha$-learnable} if it is $\alpha$-learnable
by a proper algorithm; namely an algorithm that always outputs~$q\in \Q$.


%
%

\paragraph{Distances vectors and sets.}

Let $\Q=\{q_1,\ldots,q_n\}\subseteq \Delta(\X)$,
and let $p$ be a distribution.
The $\tv$-distance vector of $p$ relative to the $q_i$'s is the vector~$v(p)=v_\Q(p)=(\tv(p,q_i))_{i=1}^n$.

Following~\cite{BKM19}, our algorithm is based on the next claim which shows that 
        in order to find $q$ such that $\tv(q,p)\leq 2\min_{i}\tv(q_i,p)+\eps$
        it suffices to find $q$ such that $v(q)\leq v(p)+\eps\cdot1_n$.

\begin{lemma}\label{lem:termination}
Let $q,p$ such that $v(q)\leq v(p) + \eps \cdot 1_n$. Then $\tv(q,p)\leq 2\min_{i} \tv(q_i,p) + \eps$.
\end{lemma}
\begin{proof}
Follows directly by the triangle inequality; indeed, let $q_i$ be a minimizer of $\tv(\cdot,p)$ in $\Q$. Then,
$\tv(q,p)\leq \tv(q,q_i) + \tv(q_i,p) \leq (\tv(p,q_i) + \eps) + \tv(q_i,p) = 2\tv(q_i,p) + \eps$.
\end{proof}

Next, we explore which $v\in\R^n$ are of the form $v=v(p)$ for some $p\in\Delta(\X)$. 
For this we make the following definition.
A vector $v \in \R^n$ is called a {\em $\tv$-distance dominating} vector if $v\geq v(p)$ for some distribution $p$. 
Define $\cP_\Q$ to be the set of all dominating distance vectors.
\begin{claim}\label{c:convexup}
$\cP_{\Q}$ is convex and upward-closed\footnote{Recall that upwards-closed means that whenever $v\in \Q_{\F}$ and $u\geq v$ then also $u\in \Q_{\F}$.}.
\end{claim}
\begin{proof}
That $\cP_{\Q}$ is upward-closed is trivial.
Convexity follows since $\tv(\cdot,\cdot)$ is convex in both of its arguments.
\end{proof}

\subsection{Pythagorian Theorem for $\kl$}

We will use the following Pythagorian theorem for the $\kl$ divergence, the version here is taken from \cite{PW15}.
\begin{lemma} \label{lem:Pythagorean}
Let $\cX$ be a set, let $\cE \subseteq \Delta(\cX)$ be a convex set of distributions, and let $p \in \Delta(\cX)$ be a distribution. Let $q^* = \arg\min_{q \in \cE} \{ \kl(q,p) \}$. Then, for all $q \in \cE$ it holds that 
$$\kl(q,p) \geq \kl(q,q^*) + \kl(q^*,p).$$
\end{lemma}

\begin{proof}
If $\kl(q,p) = \infty$, then we are done. So, we can assume $\kl(q,p) < \infty$, which also implies that $\kl(q^*,p) < \infty$. 
For $\theta \in [0,1]$, form the convex combination $q^{(\theta)} = (1-\theta) q^* + \theta q$. Since $q^*$ is the minimizer of $\kl(q,p)$, then 
$$0 \leq \left. \frac{\partial}{\partial \theta} \right\vert_{\theta = 0} \kl(q^{(\theta)},p) = \kl(q,p) - \kl(q,q^*) - \kl(q^*,p),$$
\end{proof}

If we view the picture above in the Euclidean setting, the ``triangle'' formed by $p$, $q^*$ and $q$ (for $q^*,q$ in a convex set, $p$ is outside the set) is always obtuse, and is a right triangle
only when the convex set has a ``flat face''. In this sense, the divergence is similar to the squared
Euclidean distance, and the above theorem is sometimes called the Pythagorean theorem.

\paragraph{An assumption.} 
{Our analysis uses the Minimax Theorem for zero-sum games~\cite{Neumann1928} for the same purpose that it was used in~\cite{BKM19}. 
	Therefore, we will assume a setting ({\em i.e.},\ the domain~$\X$ and the class of distributions $\Q$) in which this theorem is valid. 
	Alternatively, one could state explicit assumptions such as finiteness of $\X$ or forms of compactness under which it is known that the Minimax Theorem holds. 
	However, we believe that the presentation benefits from avoiding such explicit technical assumptions and simply assuming the Minimax Theorem 
	as an ``axiom'' in the discussed setting.}

\section{A Geometric Game from Hypothesis Selection} \label{sec:primal}

We next describe a geometric game, called the {\em ($\cP,\eps$)-primal game}. This game is between a player and an adversary, where $\cP \subseteq [0,1]^n$ 
is a given upwards-closed and nonempty convex body, and $\eps \geq 0$ is a margin parameter. 
Both~$\cP$ and $\eps$ are known to both the player and the adversary. The game proceeds in rounds roughly as follows: the player starts at position $u_0 = \vec 0 \in [0,1]^n$ and its goal is to get sufficiently close to $\cP$ as fast as possible. Let $u_k$ denote the position of the player in round $k$; if $u_k+ \eps\cdot 1_n\in \cP$ then the player wins the game.
Else, the player picks a tangent hyperplane to $\cP$ which separates $u_k+\eps\cdot 1_n$ from $\cP$ 
(such a hyperplane must exist since $u_k+ \eps\cdot 1_n\notin \cP$ ), announces it to the adversary, 
and the adversary picks the player's next position $u_{k+1}$ to be any point such that $u_{k+1}\geq u_k$ and $u_{k+1}$ is $\eps/2$-close to the tangent hyperplane chosen by the player.
The ($\cP,\eps$)-primal game is formally described in \cref{alg:game}. It uses the following notation: 
\[
\cH_{\cP,\eps}(u) = \bigl\{h\in \Delta_n : h\cdot(u+\eps\cdot 1_n) = h \cdot u + \eps < \min_{p\in \cP} \{h\cdot p\}\bigr\}.
\]
In words, $\cH_{\cP,\eps}(u)$ is the set of normals $h \in \Delta_n$ to hyperplanes separating $u + \eps\cdot 1_n$ from~$\cP$.
Note that the assumption $h\in \Delta_n$ does not lose generality, 
because $\cP$ is upwards-closed and therefore for any $u\in [0,1]^n$, $u\notin \cP$, any hyperplane separating $u$ and~$\cP$ has a normal of this form.
(See Claim 5 in \cite{BKM19} for a proof of this fact.)
Thus, by the hyperplane separation theorem, $\cH_{\cP,\eps}(u_k) = \emptyset$ if and only if $u_k+\eps\cdot 1_n\in \cP$. 
Also observe that since $\cP$ is a convex, the set $\cH_{\cP,\eps}(u)$ is convex for every $u\in\mathbb{R}^n$.

\begin{figure}
\begin{tcolorbox}
\begin{center}
{\bf The $(\cP,\eps)$-Primal Game}\\
\end{center}
\noindent

Let $\cP\subseteq [0,1]^n$ be a nonempty convex set which is upward closed.
\begin{enumerate}
\item Set $k=0$ and $u_0 = \vec 0$. 

\item While $u_k + \eps\cdot 1_n\notin \cP$ (equivalently $\cH_{\cP,\eps}(u_k) \neq \emptyset$)
\begin{enumerate}
\item The player picks a normal $h_k \in \cH_{\cP,\eps}(u_k)$ to a hyperplane tangent to $\cP$ which separates $u_k+\eps\cdot 1_n$ from $\cP$, and announces it to the adversary.

\item The adversary replies with a point $u_{k+1}$ whose every coordinate is at least as great as that of $u_k$ and is
 $\eps/2$-close to the hyperplane tangent to $\cP$ whose normal is $h_k$, {\em i.e.},
\begin{equation}\label{eq:primalrule}
u_{k+1} \geq u_k  \mbox{\hspace{3mm} and \hspace{3mm}}  h_k \cdot u_{k+1} \geq \min_{p\in \cP} \{h_k\cdot p\} - \eps/2.
\end{equation}
\item Set $k = k+1$.
\end{enumerate}
\end{enumerate}
\end{tcolorbox}
\caption{The Primal Game.}
\label{alg:game}
\end{figure}

\paragraph{Winning Strategies.}
Let $\player$ be a strategy\footnote{That is, in every round $k$, the strategy $\player$ provides a rule for picking $h_k\in \cH_{\cP,\eps}(u_k)$.} 
        for the player in the $(\cP,\eps)$-primal game. 
        A sequence $\vec 0 = u_0\leq u_1\leq \ldots\leq  u_t$ is a sequence of {\it legal-adversary moves with respect to $\player$} 
        if for every $k<t$,
        \begin{itemize}
        \item $\cH_{\cP,\eps}(u_k)\neq\emptyset$ and 
        \item $h_k \cdot u_{k+1} \geq \min_{p\in \cP} \{h_k\cdot p\} - \eps/2$, 
        where $h_k=h_k(u_k; u_{<k}, h_{<k})\in \cH_{\cP,\eps}(u_k)$ is the normal picked by $\player$ in round $k$.
        \end{itemize}
        We say that the strategy $\player$ \underline{\it wins the $(\cP,\eps)$-primal game in at most $r$ rounds} 
        if no adversary can force the game to last more than $r$ rounds. 
        That is, for every sequence $\vec 0 = u_0\leq u_1\leq \ldots\leq  u_t$ of legal adversary-moves with respect to $\player$,
        \[u_t+ \eps\cdot 1_n\notin \cP \implies t<r.\]  
%

Similarly, let $\adv$ be a strategy\footnote{That is, in every round $k$, the strategy $\adv$ provides a rule for picking $u_{k+1}$ that satisfies \Cref{eq:primalrule}.} 
        for the adversary
        in the $(\cP,\eps)$-primal game.
        A sequence $h_0,\ldots, h_{t-1}\in \Delta_n$ is a sequence of legal-player moves with respect to $\adv$
        if for every $k <t$,  $h_k\in \cH_{\cP,\eps}(u_k)\neq\emptyset$, where $u_k=u_k(h_{k-1}; u_{<k-1}, h_{<k-1})\geq u_{k-1}$ is the point picked by $\adv$ in round $k$.
        We say that the strategy $\adv$ \underline{\it forces the $(\cP,\eps)$-primal game to last at least $r$ rounds} 
        if for every sequence $h_1,\ldots, h_{t-1}\in \Delta_n$ of legal-player moves with respect to $\adv$,
        \[u_t + 1_n\cdot\eps \in \cP \implies t\geq r.\]



\subsection{Reducing Hypothesis Selection to the Primal Game}

For all that follows, we fix a finite class of distributions $\Q = \{q_1,\ldots,q_n \}$ and $\eps > 0$, and use the notation $v(\cdot)=v_Q(\cdot)$.
We next show that if the $(\cP_{\Q},\eps)$-primal game is solvable in few rounds, then $\Q$ is $2$-learnable with low sample complexity. 
The following lemma is implicitly proved in~\cite{BKM19}:

\begin{lemma} \label{lem:reduction} 
If there exists a strategy $\player$ that wins the ($\cP_{\Q},\eps$)-primal game in at most $r$ rounds, 
then $\Q$ is $2$-learnable with sample complexity $r'(\eps,\delta) = O(r\cdot \frac{\log n + \log r + \log(1/\delta)}{\eps^2}$).
\end{lemma}

The reduction is described in \cref{alg:reduction}.
It is based on \cref{lem:termination} and computes the output distribution $q$ by finding $v\in \cP_{\Q}$ such that $v\leq v(p) + \eps\cdot n$.

The following lemma is the crux of the reduction.
It is used to show that the adversary induced by the algorithm is a valid adversary for the $(\cP_\Q,\eps)$-primal game,
and provides a bound on the number of samples from $p$ which are required to compute the adversary's move.

\begin{lemma}\label{lem:progress}
Let $p \in \Delta(\cX)$ and let $\alpha,\beta>0$. Then, given $m=O(\frac{\log n + \log(1/\beta)}{\alpha^2})$ independent samples from an unknown distribution~$p$ and $h\in \Delta_n$ as an input, one can output a point $z \in \unitint^n$ that satisfies the following with probability $\geq 1-\beta$:
\begin{enumerate}
\item $h \cdot z \geq  \min_{v \in \cP_{\Q}} \{h\cdot v\}- \alpha$. 
\item $z \leq v(p)$. 
\end{enumerate}
\end{lemma}
In words, this lemma provides a procedure that, given a hyperplane tangent to $\cP_{\Q}$ and~$m$ samples from the target distribution $p$,
outputs a point $z\leq v(p)$ which is $\alpha$-close to the tangent.

\begin{proof}[Proof of \cref{lem:progress}]
By the Minmax Theorem~\cite{Neumann1928}:
\begin{align*}
\min_{v \in \cP_{\Q}} \{h\cdot v\} &= \min_{p' \in \Delta(\X)} \sum_{i \in [n]} h_i \cdot v(p')_i \tag{By definition of $\cP_{\cQ}$.}\\
&= \min_{p' \in \Delta(\X)} \sum_{i \in [n]} h_i \cdot \tv(p',q_i) \tag{By definition of $v(\cdot )$}\\
&= \min_{p' \in \Delta(\X)}\sum_{i \in [n]} h_i \max_{f_i:\X\to[0,1]} \{\Ex_{p'}[f_i] - \Ex_{q_i}[f_i]\} \tag{By definition of $\tv(\cdot,\cdot)$.}\\
&= \min_{p' \in \Delta(\X)}\max_{f_i:\X\to[0,1]}\sum_{i \in [n]} h_i(\Ex_{p'}[f_i] - \Ex_{q_i}[f_i])\\
&= \max_{f_i:\X\to[0,1]}\min_{p' \in\Delta(\X)}\sum_{i \in [n]} h_i(\Ex_{p'}[f_i] - \Ex_{q_i}[f_i])\,. \tag{By the Minmax Theorem.}
\end{align*}
Let $F_i$ for $i \in [n]$ be maximizers of the last expression. 
That is, $$(F_1,\ldots,F_n) = \argmax_{(f_1,\ldots,f_n)} \min_{p' \in \Delta(\X)}\sum_{i\in[n]} h_i(\Ex_{p'}[f_i] - \Ex_{q_i}[f_i]).$$
        By the above derivation:
\begin{equation}\label{eq:1}
 \min_{p' \in\Delta(\X)}\sum_{i \in [n]} h_i(\Ex_{p'}[F_i] - \Ex_{q_i}[F_i])= \min_{v \in \cP_{\Q}} \{h\cdot v\}.
\end{equation}
Note that the $F_i$'s depend only on the class $\cQ$ and the direction $h$; in particular they do not depend on $p$.
        Thus, the $F_i$'s can be computed by the algorithm.
        Define the point $w \in \unitint^n$ by 
\[w_i = \Ex_p[F_i] - \Ex_{q_i}[F_i],\]
and observe that $w$ can be approximated given samples from $p$.
Note that $w$ satisfies:
\begin{enumerate}
\item $\sum_{i\in [n]} h_i w_i\geq \min_{v \in \cP_{\Q}} \{h\cdot v\}  $. (By \Cref{eq:1}.)
\item $w_i = \Ex_p[F_i] - \Ex_{q_i}[F_i]\leq \max_{f_i:\X\to[0,1]} \{ \Ex_p[f_i] - \Ex_{q_i}[f_i] \} = \tv(p,q_i) = v(p)_i$.
\end{enumerate}

Thus, it suffices to output a point $z$ such that $w\geq z \geq w-\alpha\cdot 1_n$.
        This can be done using the $m=O(\frac{\log n + \log(1/\beta)}{\alpha^2})$ samples from $p$ as follows:
        use the samples to approximate $\Ex_p[F_i]$. That is, let
        \[\Ex_{\hat p} [F_i] = \frac{1}{m}\sum_{j=1}^m F_i(x_j),\]
        where $x_1,\ldots, x_m$ are the $m$ independent samples drawn from $p$.
        By a Chernoff and union bounds, we have $\lvert \Ex_{\hat p} [F_i] - \Ex_{p} [F_i]\rvert \leq \alpha/2$,
        simultaneously for all $i\leq n$.
        Therefore, the estimates $\hat z_i= \Ex_{\hat p} [F_i] - \Ex_{q_i} [F_i]$
        satisfy $\hat z_i \in (w_i-\frac{\alpha}{2}, w_i + \frac{\alpha}{2})$.
        Then, the desired vector $z$ can be taken to be $z = \hat w - \frac{\alpha}{2}\cdot 1_n$.

%
%
\end{proof}

With \cref{lem:termination}, we are ready to prove \cref{lem:reduction} which shows how to use a black-box strategy for the player 
        in the primal game to get a $2$-approximation algorithm.

\begin{proof}[Proof of \cref{lem:reduction}]
Let $\player$ be a strategy for the player that wins the ($\cP_{\Q},\eps$)-primal game in $r$ rounds. 
We will show that $\Q$ is $2$-learnable with $r' = O(r\cdot \frac{\log n + \log r + \log(1/\delta)}{\eps^2})$ samples. Let $p \in \Delta(\cX)$ be the target distribution.  The approach we use for deriving the learning algorithm is based on \cref{lem:termination} by which it suffices to find  a distribution $q \in \Delta(\cX)$ such that $v(q) \leq v(p) + \eps \cdot 1_n$. Observe that if we find $v \in \cP_{\Q}$ such that $v \leq v(p) + \eps \cdot 1_n$, a distribution $q \in \Delta(\cX)$ such that $v(q) \leq v$ can be found.

Consider the  algorithm for computing such $v$ which is depicted in Figure~\ref{alg:reduction}. 
        The algorithm is based on an execution of the primal game, where the player runs the  strategy $\player$ (see \cref{algo-line:reduction-player}) 
        and the adversary moves are based on \cref{lem:progress} (see \cref{algo-line:reduction-adverary}).

First note that \cref{lem:progress} is applies with confidence parameter $\beta=\delta/r$ and error parameter $\eps/2$.
This implies that: 
(i) the total number of samples used by the algorithm is $r$ times the sample complexity bound stated in \cref{lem:progress}
with $\alpha=\eps/2,\beta=\delta/r$, which yields the stated bound on $r'$.
(ii) With probability at least $1-\delta$, the points $z_k$ satisfy the guarantee in \cref{lem:progress} for all $k\leq r$.
In the remainder of the proof we condition on this event.
\begin{figure}
\begin{tcolorbox}
\begin{center}
{\bf A Hypothesis Selection Algorithm from the Primal Game.}\\
\end{center}
Define the set $\cP$ in the Primal Game to be $\cP_\cQ$. (The Primal Game is described in \cref{alg:game})
\begin{enumerate}
\item Set $k=0$ and $u_0 = \vec 0$. 
\item \label{algo-line:reduction-while} While $\cH_{\cP_{\Q},\eps}(u_k) \neq \emptyset$
\begin{enumerate}
\item  \label{algo-line:reduction-player} Run $\player$ to get $h_k=h_k(u_k;u_{<k},h_{<k}) \in \cH_{\cP_{\Q},\eps}(u_k)$.
\item  \label{algo-line:reduction-adverary} Let  $z_k$ be the point $z$ promised by \cref{lem:progress} applied with $u=u_k$, $h=h_k$, $p = p$, $\alpha=\eps/2$ and $\beta=\delta/r$. Define $u_{k+1}$ by setting $(u_{k+1})_i = \max \{ (z_k)_i, (u_k)_i \}$ for all~$i \in [n]$. 
\item Set $k = k+1$.
\end{enumerate}
\item Output $v = u_k + \eps \cdot 1_n$.
\end{enumerate}
\end{tcolorbox}
\caption{A Hypothesis Selection Algorithm from the Primal Game.}
\label{alg:reduction}
\end{figure}

We next claim that the adversary strategy given in \cref{algo-line:reduction-adverary} provides a sequence of legal-adversary moves w.r.t $\player$. 
To this end, we need to show that the point $u_{k+1}$ satisfies $u_{k+1} \geq u_k$ and $h_k \cdot u_{k+1} \geq \min_{p\in \cP} \{h_k\cdot p\} - \eps/2$. The former is obvious from the definition of $u_{k+1}$ in \cref{algo-line:reduction-adverary}. The latter follows since
\begin{align*}
h_k \cdot u_{k+1} &\geq h_k\cdot z_k \tag{$u_{k+1} \geq z_k, h_k\geq 0$}\\
                           &\geq \min_{p\in \cP} \{h_k\cdot p\} - \eps/2 \tag{\cref{lem:progress}}
\end{align*}
Thus, since $\player$ wins after at most $r$ rounds, the while loop in \cref{algo-line:reduction-while} must terminate in round $t\leq r$ 
and the output $v$ satisfies $v\in \cP_{\Q}$.

Finally, it remains to show that $v \leq v(p) + \eps \cdot 1_n$.  We  show that $u_k \leq v(p)$ for every $k$ by induction on $k$ (this implies $v =  u_{t} + \eps \cdot 1_n \leq v(p) + \eps \cdot 1_n $): For $k=0$, it is clearly the case that $u_0 = \vec 0 \leq v(p)$. Assume that the claim holds for some $k$ and prove it for $k+1$. By the second item in \cref{lem:progress}, $z_k \leq v(p)$, and by the induction hypothesis, $u_k \leq v(p)$. This implies that $u_{k+1} \leq v(p)$.


\end{proof}

\section{A Dual Game: Cutting-With-Margin}
One of the key steps in our solution is to adapt a dual point of view, where the separators/directions~$h_k$ are thought of as points in the dual space.
We next describe a second geometric game, called the {\em ($\cH,\eps$)-cutting-with-margin game} (in short, $(\cH_,\eps)$-cutting game), 
which can be seen as a manifestation of the primal game as seen in the dual space. 
This game too is between a player and an adversary, where $\cH \subseteq \Delta^n$ is a given convex body, and $\eps \geq 0$ is a margin parameter. Both~$\cH$ and $\eps$ are known to both the player and the adversary. 

The dual game proceeds in rounds roughly as follows: at the beginning, the universe is the set $\cH_0 = \cH$. In round $k$, the player chooses a point $h_k \in \cH_k$. 
The adversary then restricts the universe to a set $\cH_{k+1}$, which must to be a convex subset of $\cH_k$ that is disjoint from $B_1(h_k,\eps)$, an $\ell_1$ ball around $h_k$ with radius $\eps$.
If the new universe $\cH_{k+1}$ is not empty,  the game continues to the next round. Else, the game ends. A formal description of the dual game is given in \cref{algo:dual-game}.

\paragraph{Winning Strategies.}
Let $\player^\star$ be a strategy\footnote{That is, in every round $k$, the strategy $\player$ provides a rule for picking $h_k\in \cH_{k}$.} 
        for the player in the $(\cH,\eps)$-cutting game. 
        A sequence $\cH = \cH_0\supseteq  \cH_1\supseteq \ldots\supseteq  \cH_t$ is a sequence of {\it legal-adversary moves with respect to $\player^\star$} 
        if for every $k<t$,
        \begin{itemize}
        \item $\cH_{k}\neq\emptyset$ and 
        \item $\cH_{k+1} \cap B_1(h_k, \eps) = \emptyset$,
        where $h_k=h_k(\cH_k; \cH_{<k}, h_{<k})\in \cH_{k}$ is the point picked by $\player^\star$ in round $k$.
        \end{itemize}
        We say that the strategy $\player^\star$ \underline{\it wins the $(\cP,\eps)$-cutting game in at most $r$ rounds} 
        if no adversary can force the game to last more than $r$ rounds. 
        That is, for every sequence $\cH = \cH_0\supseteq  \cH_1\supseteq \ldots\supseteq  \cH_t$ of legal adversary-moves with respect to $\player^\star$,
        \[\cH_t\neq\emptyset \implies t<r.\]    
%

Similarly, let $\adv^\star$ be a strategy\footnote{That is, in every round $k$, the strategy $\adv$ provides a rule for picking $\cH_{k+1}$ as in \cref{algo-line:dual-game-adversary}.} 
        for the adversary in the $(\cP,\eps)$-primal game.
        A sequence $h_0,\ldots, h_{t-1}\in \Delta_n$ is a sequence of legal-player moves with respect to $\adv^\star$
        if $h_k\in \cH_{k}$ for every $k<t$, where $\cH_k=h_k(\cH_{k-1}; \cH_{<k-1}, h_{<k-1})$ is the set picked by $\adv^\star$ in round $k-1$.
        We say that the strategy $\adv^\star$ \underline{\it forces the $(\cH,\eps)$-cutting game to last at least $r$ rounds} 
        if for every sequence $h_1,\ldots, h_{t-1}\in \Delta_n$ of legal-player moves with respect to $\adv^\star$,
        \[\cH_t = \emptyset \implies t\geq r.\]


\begin{figure}
\begin{tcolorbox}
\begin{center}
{\bf The ($\cH,\eps$)-Cutting-With-Margin Game}\\
\end{center}
\noindent
Let $\cH\subseteq \Delta^n$ be convex.

\begin{enumerate}
\item Set $k = 0$ and $\cH_0 = \cH$. 

\item While $\cH_k \neq \emptyset$
\begin{enumerate}
\item The player picks a point $h_k \in \cH_k$  and announces it to the adversary.
\item \label{algo-line:dual-game-adversary}The adversary picks a convex set $\cH_{k+1} \subseteq \cH_k$ such that $\cH_{k+1} \cap B_1(h_k, \eps) = \emptyset$. (Observe that $\cH_{k+1} = \emptyset$ always satisfies the above conditions.) 
\item Set $k = k+1$.
\end{enumerate}
\end{enumerate}
\end{tcolorbox}
\caption{The Dual Game.}
\label{algo:dual-game}
\end{figure}

\subsection{Reduction from the Primal Game}\label{sec:reductionprimaldual}

For an upwards-closed convex set $\cP\subseteq [0,1]^n$ and $\eps>0$, 
let 
\[\cH_\cP = \cH_{\cP,\eps}({\vec 0}) = \bigl\{h\in \Delta_n :  h\cdot \vec 0  + \eps = \eps < \min_{u\in\cP}\{h\cdot u\} \bigr\}.\] 
We next show that the round complexity of the  ($\cP,\eps$)-primal game is at most the round complexity of the $(\cH_\cP,\eps)$-cutting game. 

\begin{lemma}\label{lem:reduction-primal-dual}
Let $\cP\subseteq [0,1]^n$ be an upwards-closed convex set and let $\eps\geq 0$. 
If there exists a strategy $\player^\star$ that wins the $(\cH_\cP,\frac{\eps}{4})$-cutting game in at most $r$ rounds, then there is
a strategy $\player$ that wins the ($\cP,\eps$)-primal game in at most $r$ rounds. 
\end{lemma}

The proof of \cref{lem:reduction-primal-dual} uses the following lemma.
Recall that $\cH_{\cP,\eps}(u')=\{h\in \Delta_n : h\cdot u' + \eps < \min_{u\in \cP}h\cdot u\}$.

\begin{lemma} \label{lemma:H-cap-B-empty}
Let $\cP \subseteq \unitint^n$ be an upwards-closed convex set and let  $h \in \Delta_n$, $u \in \unitint^n$.
Then,
\[h\notin \cH_{\cP,\frac{\eps}{2}}(u) \implies  \cH_{\cP,\eps}(u) \cap B_1(h,\frac{\eps}{4}) =\emptyset.\]

In other words if  $h \cdot u \geq \min_{p\in \cP} \{h \cdot p\} - \frac{\eps}{2}$ 
then, $\cH_{\cP,\eps}(u) \cap B_1(h,\frac{\eps}{4}) =\emptyset$.
\end{lemma}

\begin{proof}
Let $h\notin \cH_{\cP,\frac{\eps}{2}}(u)$, we need to show that $\cH_{\cP,\eps}(u) \cap B_1(h,\frac{\eps}{4}) =\emptyset$.
Define $G:\Delta_n \to \R$ by
\[G(h) = \min_{p\in \cP} \{h\cdot p\} - h\cdot u.\]
Thus, $G(h)$ measures the distance between $u$ and the hyperplane
tangent to $\cP$ with normal $h$.
Observe that for every $\eps'\geq 0$: 
\begin{equation}\label{eq:G}
\cH_{\cP,\eps'}(u) = \{h\in \Delta_n : G(h) > \eps'\}.
\end{equation}
Note that
\begin{itemize}
\item[(i)] $G(h) \leq \frac{\eps}{2}$ (by \Cref{eq:G}, because $h\notin \cH_{\cP,\frac{\eps}{2}}(u)$).
\item[(ii)] $G$ is $1$-Lipschitz with respect to $\ell_1$: Let $h',h'' \in \Delta_n$ and let $p^*:=\arg\min_{p\in\cP} \{h''\cdot p\}$. It holds that
\begin{align*}
 G(h')- G(h'') &=  \min_{p\in \cP} \{h'\cdot p\} - \min_{p\in \cP} \{h''\cdot p\}  -  (h'-h'') \cdot u \\
                                        &\leq h'\cdot p^* - h''\cdot p^* -  (h'-h'')\cdot u \\
                                        &= (h'-h'')\cdot (p^*-  u)\\
                                        &\leq \|h'-h''\|_1 \cdot \|p^*-u\|_\infty \tag{H\"older's inequality, $(\forall v,v') : v\cdot v' \leq \|v\|_1\|v'\|_\infty$}\\
                                        &\leq \|h'-h''\|_1. \tag{$\|p^*\|_\infty\leq 1,\, \|u\|_\infty\leq 1,\, p^*, u\geq 0$}
\end{align*}
Let $h' \in B_1(h,\frac{\eps}{4})$, so $\|h'-h\|_1 \leq \frac{\eps}{4}$. 
        Thus,
\begin{align*}
G(h') &\leq G(h) + \frac{\eps}{4} \tag{By Item (ii).}\\
        &\leq  \eps \tag{By Item (i).} 
\end{align*}
        Thus, by $\Cref{eq:G}$ $h'\notin \cH_{\cP,\eps}(u)$,
        and $\cH_{\cP,\eps}(u)  \cap B_1(h,\frac{\eps}{2}) = \emptyset$, as required.
\end{itemize}
\end{proof}

%

\begin{proof}[Proof of \cref{lem:reduction-primal-dual}]
Let $\player^\star(\cH)$ be a strategy for the player that solves the $(\cH_\cP,\frac{\eps}{4}$)-cutting game in $r$ rounds. 
        Consider the reduction described in \Cref{alg:dualtoprimalreduction} and the strategy $\player$ for the $(\cP,\eps)$-primal game 
        which is described in \Cref{item:primalstrategy}.
        Our goal is to show that $\player$ wins the game in at most $r$ rounds.
        That is, let $\vec 0 = u_0, \ldots, u_t$ be a sequence of legal-adversary moves w.r.t $\player$
        such that $u_t + \eps\cdot 1_n\notin \cP$.
        We need to show that $t < r$.
        To this end it suffices to show that the sequence $\{\cH_k\}_{k=0}^{t}$, defined in \Cref{item:primaladv} 
        is a sequence of legal-adversary moves w.r.t $\player^\star$ and that $\cH_t\neq \emptyset$.
        Indeed, by definition $\cH_0=\cH_{\cP,\eps}(\vec 0) = \cH_{\cP,\eps}$.
        Next, 
        \[\cH_{k+1} = \cH_{\cP,\eps}(u_{k+1}) \subseteq \cH_{\cP,\eps}(u_k) = \cH_k,\]
        because $u_{k+1}\geq u_k$ and $\cH_{k+1},\cH_k\subseteq \Delta_n$ contains only nonnegative vectors.
        The last property we need to show in order to establish that the $\cH_k$'s
        form a sequence of legal-adversary moves is that $\cH_{k+1} \cap B_1(h_k, \eps) = \emptyset$ for $k<t$,
        which follows from \cref{lemma:H-cap-B-empty} because $h_k \cdot u_{k+1} \geq \min_{p\in \cP} \{h_k\cdot p\} - \eps/2$
        ({\em i.e.},\ $h_k\notin \cH_{\cP,\frac{\eps}{2}}$).
        Finally, it remains to show that $\cH_t\neq \emptyset$.
        Indeed, by assumption, $u_t + \eps\cdot 1_n\notin \cP$ and therefore there must be a hyperplane separating $u_t+\eps\cdot 1_n$
        from~$\cP$. Hence, the normal to this hyperplane (normalized so that it is in $\Delta_n$) belongs to $\cH_{\cP,\eps}(u_t)=\cH_t$ and witnesses $\cH_t\neq\emptyset$.

\end{proof}

\begin{figure}
\begin{tcolorbox}
\begin{center}
{\bf Dual Player Strategy $\implies$ Primal Player Strategy.}\\
\end{center}
Let $\cP\subseteq [0,1]^n$ be an upward-closed convex set, let $\eps>0$, and consider the $(\cP,\eps)$-primal Game.
Let $\player^\star(\cdot)$ be a strategy for the player in the $(\cH_{\cP,\eps},\eps/4)$-cutting game.
\begin{enumerate}
\item Set $k=0$ and $u_0 = \vec 0$. 

\item While $u_k + \eps\cdot 1_n\notin \cP$ (equivalently $\cH_{\cP,\eps}(u_k) \neq \emptyset$)
\begin{enumerate}
\item \label{item:primaladv} Let $\cH_k := \cH_{\cP,\eps}(u_k)$.
\item \label{item:primalstrategy} Run $\player^\star$ to get $h_k=h_k(\cH_k; \cH_{<k}, h_{<k}) \in \cH_{\cP,\eps}(u_k)$ and announce $h_k$ to the adversary.

\item Let $u_{k+1}$ denote the next point which is picked by the adversary.
{\em I.e.}, $u_{k+1} \geq u_k$ and $h_k \cdot u_{k+1} \geq \min_{p\in \cP} \{h_k\cdot p\} - \eps/2$.

\item Set $k = k+1$.
\end{enumerate}
\end{enumerate}
\end{tcolorbox}
\caption{A reduction which uses a black box access to a strategy $\player^*$ in the $(\cH_{\cP,\eps},\eps/4)$-cutting game
and produces a strategy $\player$ for the $(\cP,\eps)$-primal game. This reduction is used in the proof of \Cref{lem:reduction-primal-dual}.}
\label{alg:dualtoprimalreduction}
\end{figure}

\subsection{Solution for the Cutting-With-Margin Game}\label{sec:dualsolution}

\begin{theorem} \label{thm:dual-solution}
For every convex set $\cH$ and $\eps >0$, the $(\cH,\eps)$-cutting game is solvable in $O\left(\frac{\log(n)}{\eps^2}\right)$ rounds.
\end{theorem}

\begin{proof}
Consider the strategy $\player^\star$ for the player in the $(\cH,\eps)$-cutting game that is depicted in \Cref{alg:dualstrategy}.

\begin{figure}
\begin{tcolorbox}
\begin{center}
{\bf A Strategy for the Player in the Cutting-With-Margin Game.}\\
\end{center}
\begin{center}
        In each round $k$, given a universe $\cH_k$ from the adversary,
        the player outputs $h_k=\arg\max_{h'\in \cH_k}\{\bH(h')\}$.
\end{center}
\end{tcolorbox}
\caption{A strategy for the player in the $(\cH,\eps)$-cutting game. 
Recall that $\bH(\cdot)$ denotes the entropy function, and that $\cH\subseteq \Delta_n$
and therefore $\bH(\cdot)$ is defined on every~$h\in \cH$.}
\label{alg:dualstrategy}
\end{figure}

Fix a sequence $\cH=\cH_0\supseteq \ldots \supseteq \cH_t$ of legal-adversary moves w.r.t $\player^\star$ such that $\cH_t\neq\emptyset$.
Our goal is to prove that $t\leq O(\log n/\eps^2)$.

For $k\leq t$, let $h_k = \arg\max_{h\in \cH_k}\bH(h)$ denote the point chosen by $\player^\star$.
Note that $h_k$ is well defined since $\cH_k\neq\emptyset$ for $k\leq t$.
We next prove that for every $k < r$,
\begin{equation}\label{eq:suffices}
\bH(h_k) - \bH(h_{k+1}) \geq \frac{\eps^2}{8}.
\end{equation}
This implies the desired bound $t \leq O\left(\frac{\log(n)}{\eps^2}\right)$ as follows: since the entropy function satisfies $0 \leq \bH(h) \leq \log(n)$ for every $h\in \Delta_n$. 
In particular, by \cref{eq:suffices}, 
\[0 \leq \bH(h_t) \leq \bH(h_0) -  \frac{t \cdot \eps^2}{8} \leq \log(n) - \frac{t \cdot \eps^2}{8},\] 
and therefore 
$t \leq \frac{8\log(n)}{\eps^2}$, as required.

It remains to prove \Cref{eq:suffices}.
Since  $\cH=\cH_0\supseteq \ldots \supseteq \cH_t$ is a sequence of legal-adversary moves, it holds that $\cH_{k+1} \cap B_1(h_k, \frac{\eps}{2}) = \emptyset$. 
Since $h_{k+1} \in \cH_{k+1}$, it holds that $h_{k+1} \notin B_1(h_k, \frac{\eps}{2})$, and therefore, $\| h_{k+1} - h_k \|_1 \geq \frac{\eps}{2}$.
Let $u\in\Delta_n$ denote the uniform distribution $u=(\frac{1}{n},\ldots,\frac{1}{n})$.
\begin{align*}
\bH(h_k) -  \bH(h_{k+1})  &= \kl(h_{k+1},u)-\kl(h_k, u)  \\
                             &\geq \kl(h_{k+1},h_k) \tag{\cref{lem:Pythagorean}, see reasoning below}\\
                             &\geq \frac{1}{2}  \| h_{k+1} - h_k \|_1^2 \tag{Pinsker's Inequality} \\
                             &\geq \frac{\eps^2}{8}. \tag{$\| h_{k+1} - h_k \|_1 \geq \frac{\eps}{2}$}
\end{align*}
For the second transition, the inequality  $\kl(h_{k+1},u)-\kl(h_k, u) \geq \kl(h_{k+1},h_k)$ follows from the Pythagorean Theorem for the Kullback-Leibler divergence, \cref{lem:Pythagorean}, by taking $\cE = \cH_k$, $p = u$, $q^* = h_k$, $q = h_{k+1}$.
For the theorem to apply, we need to use the facts that $\cH_k$ is convex, that $h_{k} \in \cH_k$, and that $h_{k+1}\in \cH_{k+1}\subseteq \cH_k$. We also need
\[h_k = \arg\max_{h\in \cH_k}\{\bH(h)\} = \arg\min_{h\in \cH_k}\{\kl(h,u)\}.\] 
See Figure~\ref{fig:pythagorean} for an illustration.

\end{proof}

\begin{figure}
\begin{center}
\includegraphics[ height=8cm]{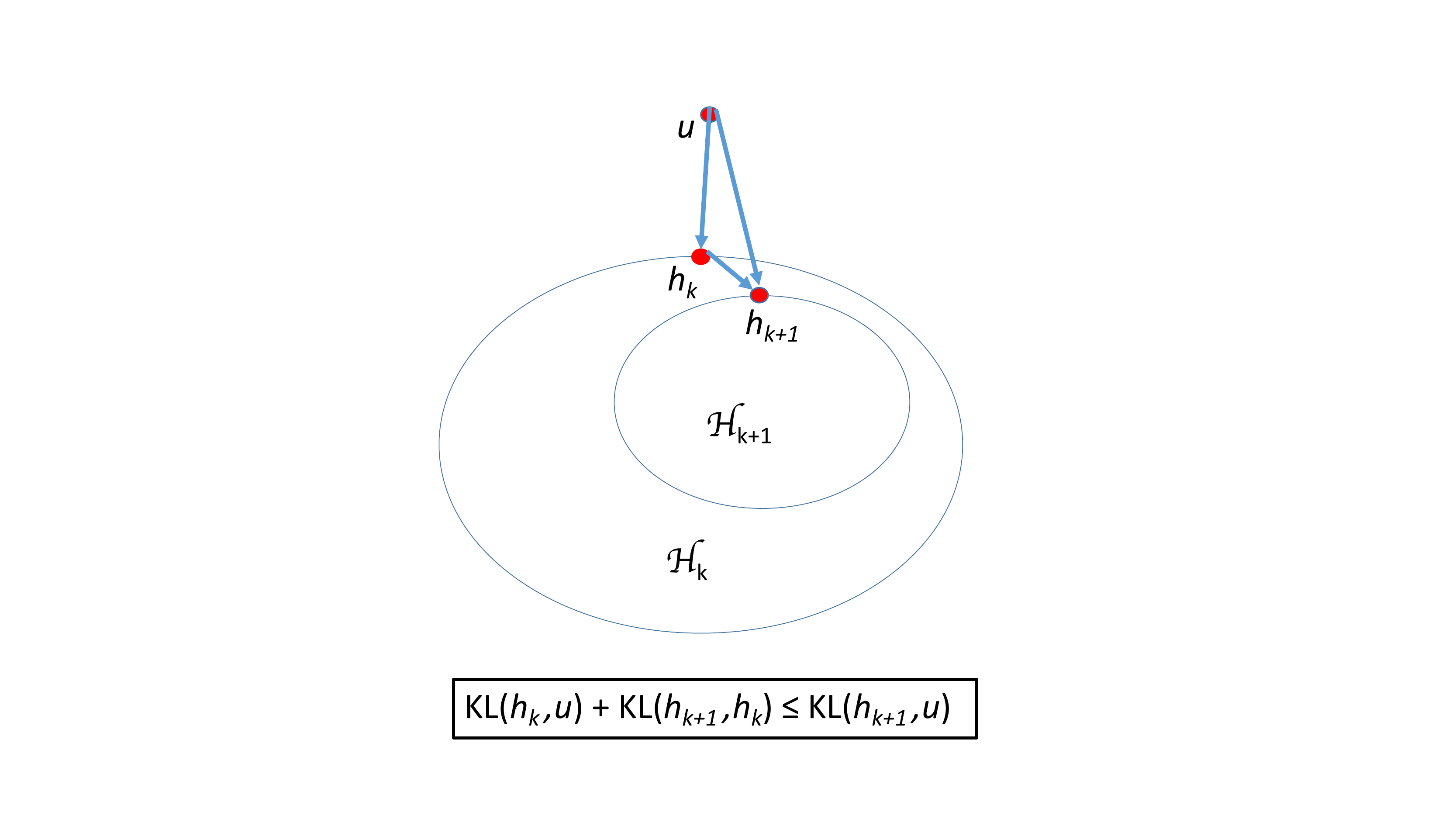}
\end{center}
\caption{An illustration of the Pythagorean Theorem for Kullback-Leibler divergence as it is used in the proof of \cref{thm:dual-solution}.} 
\label{fig:pythagorean}
\end{figure}

\subsection{Hypothesis Selection with $\poly(\log n /\eps)$ Samples}

{The results obtained so far already suffice for constructing a hypothesis selection algorithm with sample complexity of $\poly(\log n /\eps)$, as suggested by the proposition below. The algorithm we obtain in this subsection will be refined in \Cref{sec:optimal} to obtain an algorithm with near optimal sample complexity.}

\begin{proposition}
Let $\Q$ be a finite class of distributions and let $n=\lvert \Q\rvert$.
Then, $\Q$ is $2$-learnable with sample complexity 
\[m(n,\eps,\delta) = O \left( \frac{\log^2 n + \log n\log(1/\eps) + \log n \log(1/\delta)}{\eps^4} \right).\]
\end{proposition}

\begin{proof}
Let $\cP=\cP_\Q$ denote the set of all dominating-distance vectors w.r.t $\Q$.
By \cref{lem:reduction}, it suffices to give a strategy for the player that wins the $(\cP,\eps)$-primal game
in at most $r=O(\log n/\eps^2)$ rounds. The existence of such a strategy follows from \cref{thm:dual-solution},
which yields a strategy for the player that wins the $(\cH_{\cP,\eps},\eps/4)$ in $O(\log n/\eps^2)$ rounds,
and by \cref{lem:reduction-primal-dual} which transforms this strategy to a strategy that wins the $(\cP,\eps)$ game in the same number of rounds.

See \cref{alg:entirealgorithm} for a pseudo-code of the algorithm obtained by this series of reductions.
\end{proof}

\begin{figure}[h]
\begin{tcolorbox}
\begin{center}
{\bf A $2$-Approximation Algorithm for Hypothesis Selection with $\poly( \log n / \eps)$ Samples}\\
\end{center}
\noindent
Given: A class $\Q=\{q_1,\ldots,q_n\}$, and a sampling access to a target distribution $p$ and $\eps,\delta>0$.\\
Output: A distribution $p_0$ such that $\tv(p_0,p) \leq 2\min_{i}\tv(q_i,p) + \eps$ with probability at least $1-\delta$.
\begin{enumerate}
\item Let $v^*=v(p)=(\tv(p,q_i))_i\in\R^n$, and set $u_0=(0,\ldots,0)\in\R^n$. (Note that $v^*$ is not known to the algorithm)
\item For $k=1,\ldots$
\begin{enumerate}
        \item If $u_{k} + \eps\cdot 1_n \in \cP=\cP_{\Q}$ then find $q$
        such that $v(q)\leq u_k+\eps\cdot1_n$ and output it.
        \item Else, pick a separator $h_k=\arg\max_{h\in \cH_{\cP,\eps}(x_k)}\{\bH(h)\}$ with maximum entropy.
        \item    Draw $m=O(\frac{\log n + \log\log n + \log(1/\eps) + \log(1/\delta)}{\eps^2})$ samples from $p$
        to compute $u_{k+1}$ such that $u_k\leq u_{k+1}\leq v^*$, and 
        \[ h_k\cdot u_{k+1} \geq \min_{u\in \cP}\{h_k\cdot u\} - \frac{\eps}{2}.\]
        (See \cref{lem:progress} for the computation of $u_{k+1}$.)
         \item Continue to the next iteration.
\end{enumerate}
\end{enumerate}
\end{tcolorbox}
\caption{The hypothesis selection algorithm obtained by the reductions to the two games.}
\label{alg:entirealgorithm}
\end{figure}

\section{Obtaining Near Optimal Sample Complexity} \label{sec:optimal}

{In this section we prove \Cref{thm:main}, giving a $2$-approximation algorithm for hypothesis selection with sample complexity that is tight up to lower-order terms. To this end, we first study a refined version of the primal game from \Cref{sec:primal}, and then study a refined version of the hypothesis selection algorithm given in \Cref{alg:entirealgorithm} (that was based on our solution to the cutting-with-margin game).}



\begin{figure}[h]
        \begin{tcolorbox}
                \begin{center}
                        {\bf The Refined $(\cP,\eps)$-Primal Game}\\
                \end{center}
                \noindent
                
                Let $\cP\subseteq [0,1]^n$ be a nonempty convex set which is upward closed, $C_0$ here is a large constant. 
                \begin{enumerate}
                        \item Set $k=0$, $u_0 = \vec 0$, $d=1$, $d'=d/(C_0 \log (1+1/d))$. 
                        
                        \item While $d>\eps/2${\label{outerloop}}
                        \begin{enumerate}                       
                        \item While $\cH_{\cP,d-d'}(u_k) \neq \emptyset$
                        \begin{enumerate}
                                \item The player picks a normal 
                                $$h_k = \argmax_{h'\in \cH_{\cP,d-d'}(u_k)}\{\bH(h')\} \in \cH_{\cP,d-d'}(u_k)$$ to a hyperplane tangent to $\cP$ which separates $u_k+(d-d')\cdot 1_n$ from $\cP$.
                                
                                \item Run {\em Refined Hypothesis Select} algorithm\footnote{Note that stage requires $\tilde{O}(2^{2j})$ qureies} to reply with a point $u_{k+1}$, and an integer $j\in\{0,\ldots,2+\lceil \log(1+1/d) \rceil\}$  
                                which satisfy
                                \begin{equation}\label{eq:refinedprimalrule}
                                        u_{k+1} \geq u_k  \mbox{\hspace{3mm} and \hspace{3mm}} 
                                        \sum_{i} \min(2^{-j}, u_{k+1,i}-u_{k,i}) \cdot h_{k,i} > 2 d'
                                \end{equation}
                                \item Set $k = k+1$.
                        \end{enumerate}
                \item 
                 Set $d=d-d'$, $d'=d/(C_0 \log (1+1/d))$.
                \end{enumerate}
        \item Output a distribution $r$ satisfying $\tv(r,q_i)\le u_{k,i}+d$  for all $i$.
                \end{enumerate}
        \end{tcolorbox}
        \caption{The Refined Primal Game.}
        \label{alg:refgame}
\end{figure}

\subsection{Refining the Primal Game} 
{Consider the Refined Primal Game algorithm given in \Cref{alg:refgame}. This algorithm uses the Refined Hypothesis Select algorithm that can be found in \Cref{alg:refdens}.}
Let us analyze the Refined Primal Algorithm before presenting the Refined Hypothesis Select component. The critical property of the refined hypothesis select part will be that the number of samples needed to get \eqref{eq:refinedprimalrule} to hold with a given $j$ is $\tilde{O}(2^{2j})$,
which can be substantially smaller than $1/\eps^2$ required by the analogous step of the original algorithm. At the same time, when $j$ is large (and thus the sample complexity cost is large), we get a stronger estimate of 
$\Omega(2^j \cdot d')$ on the distance $\| h_{k}-h_{k+1}\|_1$, which translates into more progress towards reducing the value of $\bH(h_{k+1})$, helping the algorithm terminate faster. Thus we get a win-win situation, where small $j$ means fewer samples needed, and a large $j$ means a lot of progress towards completion. 


\begin{claim} (Refined $(\cP,\eps)$-Primal Game -- running time)
        \label{cl:ref1}
        Suppose that conditioned on outputting~$j$, {\em Refined Hypothesis Select} terminates after $\le A\cdot 2^{2j}$ samples in expectation. Then the expected number of samples needed by the Refined $(\cP,\eps)$-Primal Game
        is $\tilde{O}((A \cdot \log n)/\eps^2)$. 
\end{claim}

\begin{proof}
        Note that the outer loop (\Cref{outerloop}), where $d$ gets reduced runs a total of $O(\log^2 (1/\eps))$ times, and therefore it suffices to analyze one execution of the loop to show that as long as $d=\Omega(\eps)$, the number of samples used in reducing $d$ to $d-d'$ is bounded by $\tilde{O}((A \cdot \log n)/\eps^2)$. 
        
        Consider a single iteration of the inner loop, we would like to lower bound the difference $\bH(h_k)-\bH(h_{k+1})$. By the exection of the algorithm $\cH_{\cP,d}(u_k)=\emptyset$, and thus $h_{k+1} \notin \cH_{\cP,d}(u_k)$. Therefore, by definition of $\cH_{\cP,d}(u_k)$ it holds:
        \begin{equation}\label{eq:fail4}
        \min_{v\in \cP}\{h_{k+1} \cdot v\} \le h_{k+1} \cdot u_k + d.
\end{equation}
On the other hand, $h_{k+1}\in \cH_{\cP,d-d'}(u_{k+1})$, and thus
                \begin{equation}\label{eq:fail5}
                \min_{v\in \cP}\{h_{k+1} \cdot v\} > h_{k+1} \cdot u_{k+1}+ d-d'.
        \end{equation}
Putting equations \eqref{eq:fail4} and \eqref{eq:fail5} together, we get:
  \begin{equation*}
 h_{k+1} \cdot u_{k+1}+ d-d'<   \min_{v\in \cP}\{h_{k+1} \cdot v\}   \le h_{k+1} \cdot u_k + d.
        \end{equation*}
Thus from above, 
\begin{multline*}
    d'> h_{k+1}\cdot (u_{k+1}-u_k)\ge \sum_i \min(2^{-j}, u_{k+1,i} - u_{k,i})\cdot h_{k+1,i} = \\ 
        \sum_i \min(2^{-j}, u_{k+1,i} - u_{k,i})\cdot h_{k,i} + 
        \sum_i \min(2^{-j}, u_{k+1,i} - u_{k,i})\cdot (h_{k+1,i}-h_{k,i}).
\end{multline*}
Applying   \eqref{eq:refinedprimalrule} on the  RHS we get: 

\begin{multline*}
       d'> 2 d' +  \sum_i \min(2^{-j}, u_{k+1,i} - u_{k,i})\cdot (h_{k+1,i}-h_{k,i}) >
        2 d' - 2^{-j} \cdot \| h_{k+1}-h_k\|_1.
\end{multline*}
 Therefore,
        \begin{equation}
                \label{eq:fail6}
                 \| h_{k+1}-h_k\|_1> d' \cdot 2^j. 
        \end{equation} 
By the same derivation as in the proof of \Cref{thm:dual-solution}, \Cref{eq:fail6} implies 
\begin{align*}     \label{eq:fail7}
\bH(h_k) -  \bH(h_{k+1})  &= \kl(h_{k+1},u)-\kl(h_k, u)  \\
                             &\geq \kl(h_{k+1},h_k) \tag{\cref{lem:Pythagorean}}\\
                             &\geq \frac{1}{2}  \| h_{k+1} - h_k \|_1^2 \tag{Pinsker's Inequality} \\
                             &> \frac{d'^2}{2} \cdot 2^{2j} \tag{Equation \eqref{eq:fail6}}\\
                             &>\frac{\eps^2/ \log^2 (1/\eps)}{4}\cdot 2^{2j}=\tilde{\Omega}(\eps^2 \cdot 2^{2j}).
\end{align*}

         At the beginning of execution with a given $d$, $\bH(h_k)\le \log n$, and at the end it is at least $0$. Each step causing a reduction by $\tilde{\Omega}(\eps^2 \cdot 2^{2j})$ takes $\le A \cdot 2^{2j}$ queries. Thus the total number of queries for a given value of $d$ is bounded by $\tilde{O}((A\log n)/\eps^2$. 
\end{proof}

\begin{claim} (Refined $(\cP,\eps)$-Primal Game -- correctness)
        \label{cl:ref2}
        Let $i\in[n]$ be any fixed index, which may depend on $p$ and the $q$'s but not on the execution of the algorithm.
        Suppose that at every step $k$ of {\em Refined Hypothesis Select}, the probability $$\Pr[u_{k+1,i}>\tv(p,q_i)]<\tilde{o}(\delta\eps^2/\log n).$$ Then the probability that 
        $\tv(r,q_i)> \tv(p,q_i)+\eps/2$ is at most $\delta$. 
\end{claim}

\begin{proof}
        At each step, $\bH(h_k)$ decreases by at least $\tilde{\Omega}(\eps^2)$, and thus the total number of calls to the Refined Hypothesis Select algorithm is 
        $\tilde{O}((\log n)/\eps^2)$. Therefore, by union bound, except with probability $<\delta$, at each step $k$, $u_{k,i}\le \tv(p,q_i)$. Therefore, the distribution $r$ the algorithm outputs satisfies
        $$\tv(r,q_i)\le u_{k_{end},i}+d \le \tv(p,q_i)+d<\tv(p,q_i)+\eps/2.$$
\end{proof}

\subsection{Refining the Hypothesis Selection Algorithm}

{We next turn our attention to the Refined Hypothesis Select algorithm in \Cref{alg:refdens}.} 

\begin{figure}
        \begin{tcolorbox}
                \begin{center}
                        {\bf The Refined Hypothesis Select Algorithm}\\
                \end{center}
                \noindent       Given $d,d'=d/(C_0 \log (1+1/d))$, error parameter $\ga>0$, a point  $u$ such that  $\cH_{\cP_\cQ,d}(u)=\emptyset$, and a distribution $h\in         
                \cH_{\cP_\cQ,d-d'}(u)$ the algorithm will output $j\in\{0,\ldots,2+\lceil \log(1+1/d) \rceil\}$,  a point~$v$ and $n$ functions $F_i:\cX \rightarrow [0,1] , i=1,\ldots, n$ such that the following properties
                hold:
                \begin{enumerate}[label=(\roman*)]
                        \item {\label{prop1}}
                        The algorithm outputs `success' with probability $>1-\ga$, where the failure event only depends on the randomness of the samples the algorithm receives;
                        \item $v\ge u$;{\label{prop2}}
                        \item $ \sum_{i} \min(2^{-j}, v_i-u_i) \cdot h_i > 2d'${\label{prop3}}
                        \item For any $i\in[n]$  which is fixed in advance (unknown to the algorithm) if
                        $v_i>u_i$, then except with probability $\ga$, $\Ex_p[F_i]-\Ex_{q_i}[F_i]>v_i +2^{-j-2}$. {\label{prop4}}
                        \end{enumerate}
Algorithm:
        \begin{enumerate}
                \item 
                Let $\{F_i\}_{i=1}^n$ be as in the proof of \Cref{lem:progress}:
                $F_i:\cX\rightarrow [0,1]$ such that 
                \begin{equation}
                        \sum_{i \in [n]} h_i\cdot(\Ex_{p}[F_i] - \Ex_{q_i}[F_i])\ge \min_{v \in \cP_{\Q}} \{h\cdot v\}.
                \end{equation}
           \item  For 
           $j\in\{0,\ldots,2+\lceil \log(1+1/d) \rceil\}$:
                \begin{enumerate}
                        \item
                        Use $m_j:=C_1 \log (\log (1/d)/\ga) \cdot 2^{2j}$ samples $\{x_k\}_{k=1}^{m_j}$ from $P$ to generate empirical estimates 
                        \begin{equation}
                                w_{ji}:=\frac{1}{m_j} \sum_{k=1}^{m_j} F_i(x_k)- \Ex_{q_i}[F_i]; 
                        \end{equation}
                \item 
                   Set 
                   \begin{equation}
                        v_{ji}:= \max(u_i, w_{ji}-2^{-j-1}); 
                   \end{equation}
             \item If   $\sum_{i} \min(2^{-j}, v_{ji}-u_i) \cdot h_i >2 d'$:
             \begin{enumerate}\item  set
             $v:=v_j$ \item {\bf terminate and output $(j,v,\{F_i\})$}
             \end{enumerate}
                \end{enumerate}
        \item 
    If the loop hasn't terminated for any $j$, {\bf output `Fail' and restart the algorithm}. 
    \end{enumerate}
        \end{tcolorbox}
        \caption{The Refined Hypothesis Select Algorithm.}
        \label{alg:refdens}
\end{figure}

Note that the number of samples used by the Refined Hypothesis Select algorithm 
is spelled out explicitly. Therefore, our only task is to show that its success guarantees hold. 
Properties \ref{prop2} and \ref{prop3} holds due to stopping conditions of the algorithm. 
Next claim proves that Property \ref{prop4} holds. 
\begin{claim} \label{cl:ref4} Fix an index $i$.
        Assuming Refined Hypothesis Select algorithm does not output `Fail', 
        the probability of the event \begin{equation}\label{eq:badEvent}\Pr[(v_i>u_i)\wedge (\Ex_p[F_i]-\Ex_{q_i}[F_i]\le v_i +2^{-j-2})]<\ga.\end{equation}
\end{claim}

\begin{proof}
        Note that $v_i = \max(u_i, w_{ji}-2^{-j-1})$. Therefore, $v_i>u_i$ iff
         $w_{ji}>u_i+2^{-j-1}$. 
         Recall that $w_{ji}=  \sum_{k=1}^{m_j} F_i(x_k)- \Ex_{q_i}[F_i]$.
         Therefore event $\{v_i>u_i\}$ dominated by the event $ \left(\frac{1}{m_j} \sum_{k=1}^{m_j} F_i(x_k)- \Ex_{q_i}[F_i]=v_i+2^{-j-1}\right)$.
         Therefore, the event from the claim is equal to the event 
        $$
        (\Ex_p[F_i]-\Ex_{q_i}[F_i]\le v_i +2^{-j-2}) \wedge \left(\frac{1}{m_j} \sum_{k=1}^{m_j} F_i(x_k)- \Ex_{q_i}[F_i]=v_i+2^{-j-1}\right),
        $$
        which is dominated by the event 
        $$
        \left|\Ex_p[F_i] - \frac{1}{m_j} \sum_{k=1}^{m_j} F_i(x_k) \right| \ge 2^{-j-2}. 
        $$
        By Chernoff bound, this probability is bounded by $c_2 \ga/(\log d)$ for a small constant $c_2$ (which depends on $C_1$). By taking union bound on the different possible $j$'s in the algorithm, we obtain an upper bound of $\ga$ on the failure probability.  
\end{proof}

Next -- more importantly -- we need to establish that the probability that the algorithm outputs `Fail' is bounded by $\ga$ (First property of the algorithm). 

\begin{claim}\label{cl:ref3}
        The probability that the Refined Hypothesis Select algorithm outputs `Fail'
        is $<\ga$, where the randomness comes from the samples from $P$ that it receives. 
\end{claim}

\begin{proof}
        Our starting point is the fact that  $h\in       
        \cH_{\cP_\cQ,d-d'}(u)$, and therefore 
$       \min_{v \in \cP_{\Q}} \{h\cdot v\} > h\cdot u + d-d'$. Hence
\begin{equation*}
                \sum_{i \in [n]} h_i\cdot(\Ex_{p}[F_i] - \Ex_{q_i}[F_i]-u_i) > d-d'.
\end{equation*}
Partition the set of coordinates $[n]$ as follows. Let 
\begin{equation}{\label{Sjdef}}
S_j:=\{i\in[n]: \Ex_{p}[F_i] - \Ex_{q_i}[F_i] -u_i\in (2^{-j},2^{-j+1}]\}
\end{equation}
for  $j\in\{0,\ldots,2+\lceil \log(1+1/d) \rceil\}$. Denote $j_{max}:=2+\lceil \log(1+1/d) \rceil\}$. Note that the sets $S_j$ are mutually disjoint. Some coordinates may belong to none of the sets, but only if 
$ \Ex_{p}[F_i] - \Ex_{q_i}[F_i]- u_i<2^{-j_{max}}$. We have
\begin{multline*}
        \sum_{j=0}^{j_{max}} \sum_{i \in S_j} h_i\cdot(\Ex_{p}[F_i] - \Ex_{q_i}[F_i]-u_i)  >\\    \sum_{i \in [n]} h_i\cdot(\Ex_{p}[F_i] - \Ex_{q_i}[F_i]-u_i-2^{-j_{max}}) > d-d'-2^{-j_{max}} >\frac{d}{2}. 
\end{multline*}

Therefore, there exists some $j$ such that
 $$ \sum_{i \in S_j} h_i\cdot(\Ex_{p}[F_i] - \Ex_{q_i}[F_i]-u_i)> \frac{d}{2j_{max}} $$

Therefore, for any constant $C_2>0$, for a sufficiently large $C_0$ the is a $j$ such that 
\begin{equation}\label{eq:fail1}
\sum_{i \in S_j} h_i\cdot(\Ex_{p}[F_i] - \Ex_{q_i}[F_i]-u_i) > C_2\cdot d'. 
\end{equation}
Note that \eqref{eq:fail1} and \eqref{Sjdef} implies 
\begin{equation}\label{eq:fail2}
        \sum_{i \in S_j} h_i > C_2\cdot d'\cdot 2^j /2. 
\end{equation}
We claim that for a sufficiently large constant $C_2$, the algorithm will terminate at step $j$ with probability $>1-\ga$ (assuming it hasn't terminated earlier). Thus, the failure probability of the algorithm is bounded by $\ga$. 

For any given $i\in S_j$, we have by the Chernoff bound
$$
\Pr[w_{ji}>(\Ex_{p}[F_i] - \Ex_{q_i}[F_i]-2^{-j-2})] >1-\ga/2, 
$$
and thus 
$$
\Pr[w_{ji}>u_i+3\cdot 2^{-j-2})] >1-\ga /2. 
$$
Therefore 
$$
        \Ex\left[ \sum_{i\in S_j} h_i \cdot 1_{w_{ji}\le u_i+3\cdot 2^{-j-2}}\right]<
        \frac{\ga}{2}\cdot      \sum_{i \in S_j} h_i. 
$$
Therefore, by Markov inequality, with probability at least $1-\ga$, 
$$
\sum_{i\in S_j} h_i \cdot 1_{w_{ji}\le u_i+3\cdot 2^{-j-2}}<
\frac{1}{2}\cdot        \sum_{i \in S_j} h_i.
$$
Hence, with probability at least $1-\ga$, 
\begin{equation}\label{eq:fail3}
\sum_{i\in S_j} h_i \cdot 1_{w_{ji}> u_i+3\cdot 2^{-j-2}}>
\frac{1}{2}\cdot        \sum_{i \in S_j} h_i.
\end{equation}
We claim that assuming \eqref{eq:fail3} holds, the algorithm will terminate at step $j$. We have
\begin{multline*}
        \sum_{i} \min(2^{-j}, v_i-u_i) \cdot h_i  \ge 
        \sum_{i\in S_j:~ w_{ji}> u_i+3\cdot 2^{-j-2}}  \min(2^{-j}, v_i-u_i)\cdot h_i=\\
        \sum_{i\in S_j:~ w_{ji}> u_i+3\cdot 2^{-j-2}}  \min(2^{-j}, w_{ji}-2^{-j-1})\cdot h_i\ge \\       \sum_{i\in S_j:~ w_{ji}> u_i+3\cdot 2^{-j-2}}
        2^{-j-2} \cdot h_i \ge \frac{2^{-j-2}}{2}\cdot  \sum_{i \in S_j} h_i \ge \\
\frac{2^{-j-2}}{2}\cdot         C_2\cdot d'\cdot 2^j /2 = \frac{C_2 \cdot d'}{16} >2 d',
\end{multline*}
when $C_2>32$ -- guaranteeing that the algorithm terminates. 
\end{proof}

\Cref{cl:ref3} implies:
\begin{claim}
\label{cl:ref5} Assuming $\ga< d^3$, the expected number of queries contributed by `Fail's is an additive $O(1)$.  
\end{claim}

\begin{proof}
        The `Fail' state is reached with probability $<\ga<d^3$, while the number of queries of one run of the main loop is bounded by $\tilde{O}(1/d^2)$. 
\end{proof}

\subsection{Proof of \Cref{thm:main}}

{We are new ready to prove our main result, \Cref{thm:main}.} We break the proof into two cases: the case when $\delta$ is not too small: $\delta\ge \eps^2/n^3$, and the case when $\delta<\eps^2/n^3$ is very small. 
        
        \paragraph{The case $\delta\ge \eps^2/n^3$.} In this case, we simply run the Refined Primal Game algorithm from \Cref{alg:refgame}, where we set the error parameter $\ga$ in the Refined Hypothesis Select algorithm to 
        $\tilde{o} (\delta\eps^2/\log n)$. 
        
\paragraph{Correctness.}        By \Cref{cl:ref2} applied to $i^*:= \argmin_i \tv(p,q_i)$, we have, with probability $>1-\delta$, the output $r$ satisfies 
        $$\tv(r,q_{i^*})\le  \tv(p,q_{i^*})+\eps/2,$$
therefore,
\begin{equation}
\tv(r,p) \le \tv(r,q_{i^*}) + \tv(p,q_{i^*}) < 2\cdot  \tv(p,q_{i^*}) + \eps = 
2\cdot \min_i \tv(p,q_{i}) + \eps.
\end{equation}

\paragraph{Sample complexity.} The conditions of \Cref{cl:ref1} are met with $A=\tilde{O}(\log (1/\delta))$. Therefore, by \Cref{cl:ref1}, the total sample complexity in this case is bounded by 
\begin{equation}
        \tilde{O} \left( \frac{\log n \cdot  \log (1/\delta)}{\eps^2}\right).
\end{equation}

\paragraph{The case $\delta< \eps^2/n^3$.} Consider the algorithm on \Cref{alg:lowError}.

        \begin{figure}
                \begin{tcolorbox}
                        \begin{center}
                                {\bf Hypothesis Selection in the Tiny Error Regime}\\
                        \end{center}
                \begin{enumerate}
                        \item Repeat the following until {\bf Success} is reached:
                        \begin{enumerate}
                                \item Run the refined Primal Game with $\delta'=\eps^2/n^3$ to obtain a distribution $r$; 
                                \item Use $\tilde{O}(\log (1/\delta^2)/\eps^2)$ fresh samples to verify that except with probability $<\delta/2$, for all calls of Hypothesis Selection Algorithm, whenever $v_{ji}>u_i$, we have 
                                $\Ex_p[F_i]-\Ex_{q_i}[F_i]>v_{ji}$. 
                                \begin{enumerate}
                                        \item 
                                        if verification passes, output {\bf Success} and the distribution $r$; 
                                        \item 
                                        otherwise, restart the calculation.  
                                \end{enumerate}
                        \end{enumerate}
                        \end{enumerate}
                \end{tcolorbox}
                \caption{The Tiny Error Case}
                \label{alg:lowError}
        \end{figure}

\paragraph{Correctness.} The number of calls to Hypothesis Selection Algorithm is significantly smaller than $o(1/\delta)$. Therefore, by union bound, the probability of {\bf Success} being returned despite 
        $\Ex_p[F_i]-\Ex_{q_i}[F_i]<v_{ji}$ holding at some point of the execution is 
        $o(\delta)$. Assuming $\Ex_p[F_i]-\Ex_{q_i}[F_i]>v_{ji}$ at all steps of the execution, the algorithm outputs a correct solution. 

\paragraph{Sample complexity.} In this case, we first run the previous case with $\delta'=\eps^2/n^3$. As seen above, this step only requires 
$$      \tilde{O} \left( \frac{\log^2 n}{\eps^2}\right)
$$
samples. Moreover, as noted earlier, by union bound, with probability $>1-1/n$, the event \eqref{eq:badEvent} from \Cref{cl:ref4} never happens throughout the execution of the algorithm. When \eqref{eq:badEvent} doesn't happen, we have
$$
\Ex_p[F_i]-\Ex_{q_i}[F_i]>v_{ji}+2^{-j-2} > v_{ji}+\tilde{\Omega}(\eps),
$$
and verification will pass with probability $>1-\delta'$. 
 Therefore, the expected number of samples that will be needed until {\bf Success} is reached is bounded by a $$(1+o(1))\cdot (\text{number
        of samples used by one iteration}) = 
        \tilde{O} \left( \frac{\log^2 n+\log (1/\delta)}{\eps^2}\right).$$

\section*{Acknowledgements}
We thank Abbas Mehrabian and Hassan Zokaei Ashtiani for discussions regarding the implied improvement of our work to learning mixtures of gaussians.
We also thank Naman Agarwal, Elad Hazan, Tomer Koren, and Karan Singh for fruitful discussions concerning the connections between the cutting-with-margin game and online optimization.


\bibliographystyle{alpha}
\bibliography{distlearn}

\end{document}